\documentclass{article}

\usepackage[accepted]{icml2024}

\usepackage{appendix}
\usepackage[utf8]{inputenc} %
\usepackage[T1]{fontenc}    %
\usepackage{hyperref}       %
\usepackage{url}            %
\usepackage{booktabs}       %
\usepackage{amsfonts}       %
\usepackage{amsmath}        %
\usepackage{amssymb}        %
\usepackage{bbm}            %
\usepackage{graphicx}       %
\usepackage[acronym]{glossaries}  %
\usepackage{nicefrac}       %
\usepackage{mathtools}      %
\usepackage{microtype}      %
\usepackage{multirow}       %
\usepackage{subcaption}
\usepackage{caption}
\usepackage{xcolor}         %
\usepackage{placeins}       %
\usepackage{fix-cm}
\usepackage{enumitem}       %

\newacronym{gbp}{GBP}{Gaussian Belief Propagation}
\newacronym{ourmethod}{\acrshort*{gbp}Learn}{\acrshort*{gbp} Learning}

\newcommand{\p}{p}

\newcommand{\lrb}[1]{\left(#1\right)}
\newcommand{\lrbnorm}[1]{\left\lVert#1\right\rVert}
\newcommand{\lrbtwonorm}[1]{\lrbnorm{#1}_2}

\newcommand{\R}{\mathbb{R}}

\newcommand{\allvars}{\mathbf{X}}
\newcommand{\var}{x}
\newcommand{\varbf}{\mathbf{x}}
\newcommand{\Jac}{\mathbf{J}}
\newcommand{\allfacs}{\Phi}
\newcommand{\fac}{\phi}
\newcommand{\varbffac}[1]{\varbf_{\fac_{#1}}}  %
\newcommand{\varbffaclin}[1]{\varbf_{\fac_{#1},0}}  %
\newcommand{\nei}[1]{\mathrm{ne}\lrb{#1}}
\newcommand{\obs}{\mathbf{y}}
\newcommand{\msgnoedge}{m}
\newcommand{\msg}[2]{\msgnoedge_{{#1\rightarrow#2}}}
\newcommand{\measfn}[1]{\mathbf{h}\lrb{#1}}
\newcommand{\nconnvar}{V}

\newcommand{\Prec}{\Lambda}
\newcommand{\info}{\eta}
\newcommand{\Precmsg}[2]{\Prec_{{#1\rightarrow #2}}}
\newcommand{\infomsg}[2]{\info_{{#1\rightarrow #2}}}
\newcommand{\allPrecmsgs}{\mathbf{D}}
\newcommand{\allinfomsgs}{\mathbf{e}}
\newcommand{\Covfacplusmsg}{\mathbf{S}}
\newcommand{\robthresh}{N_{\mathrm{rob}}}

\newcommand{\filter}{\mathbf{\theta}}
\newcommand{\filters}{\mathbf{\Theta}}
\newcommand{\weight}{\mathbf{W}}
\renewcommand{\input}{\mathbf{x}}
\newcommand{\inputpatch}{\mathbf{X}}
\newcommand{\inputscal}{x}

\newcommand{\biasscal}{b}
\newcommand{\densebiasscal}{d}
\newcommand{\densebias}{\mathbf{\densebiasscal}}
\newcommand{\filtsize}{K}
\newcommand{\Econv}{E_\mathrm{conv}}
\newcommand{\EconvT}{E_\mathrm{convT}}

\newcommand{\Emaxpool}{E_\mathrm{maxpool}}
\newcommand{\Eupsample}{E_\mathrm{upsample}}

\newcommand{\Edense}{E_\mathrm{dense}}
\newcommand{\Edensesub}[1]{E_\mathrm{dense,#1}}
\newcommand{\Esm}{E_\mathrm{softmax}}

\newcommand{\inputobs}{\mathbf{z}}
\newcommand{\allparams}{\mathbf{\Psi}}
\newcommand{\layerparams}{\mathbf{\psi}}

\newcommand{\psnr}{\mathrm{PSNR}}

\newcommand{\Softmax}[2]{\mathrm{softmax}_{#1}\left(#2\right)}
\newcommand{\flatten}[1]{\mathrm{vec}\lrb{#1}}
\newcommand{\tpose}{^\top}

\DeclarePairedDelimiter\floor{\lfloor}{\rfloor}
\newcommand\andy[1]{\textcolor{green}{}}
\newcommand\seth[1]{\textcolor{blue}{}}
\newcommand\todo[1]{}%

\newcommand{\myeqref}[1]{\eqref{#1}}

\newcommand{\citenop}[1]{\citeauthor{#1},~\yrcite{#1}}

\usepackage{pifont}%

\definecolor{filtervarcol}{HTML}{485B9F}
\definecolor{coeffvarcol}{HTML}{DE4E4E}
\definecolor{pixelvarcol}{HTML}{528951}

\icmltitlerunning{Learning in Deep Factor Graphs with Gaussian Belief Propagation}

\begin{document}

\twocolumn[
\icmltitle{Learning in Deep Factor Graphs with Gaussian Belief Propagation}

\icmlsetsymbol{equal}{*}

\begin{icmlauthorlist}
\icmlauthor{Seth Nabarro}{dyson}
\icmlauthor{Mark van der Wilk}{oxford}
\icmlauthor{Andrew J. Davison}{dyson}
\end{icmlauthorlist}

\icmlaffiliation{dyson}{Dyson Robotics Lab, Imperial College London, UK}
\icmlaffiliation{oxford}{Department of Computer Science, University of Oxford, UK}
\icmlcorrespondingauthor{}{sdn09@ic.ac.uk}

\icmlkeywords{Belief Propagation, Bayesian Deep Learning, Energy Based Models, Local Learning}

\vskip 0.3in
]

\printAffiliationsAndNotice{}  %

\begin{abstract}
    We propose an approach to do learning in Gaussian factor graphs. We treat all relevant quantities (inputs, outputs, parameters, activations) as random variables in a graphical model, and view training and prediction as inference problems with different observed nodes. Our experiments show that these problems can be efficiently solved with belief propagation (BP), whose updates are inherently local, presenting exciting opportunities for distributed and asynchronous training. Our approach can be scaled to deep networks and provides a natural means to do continual learning: use the BP-estimated posterior of the current task as a prior for the next. On a video denoising task we demonstrate the benefit of learnable parameters over a classical factor graph approach and we show encouraging performance of deep factor graphs for continual image classification.
\end{abstract}

\section{Introduction}

\begin{figure}[t]
    \centering
    \includegraphics[width=0.47\textwidth]{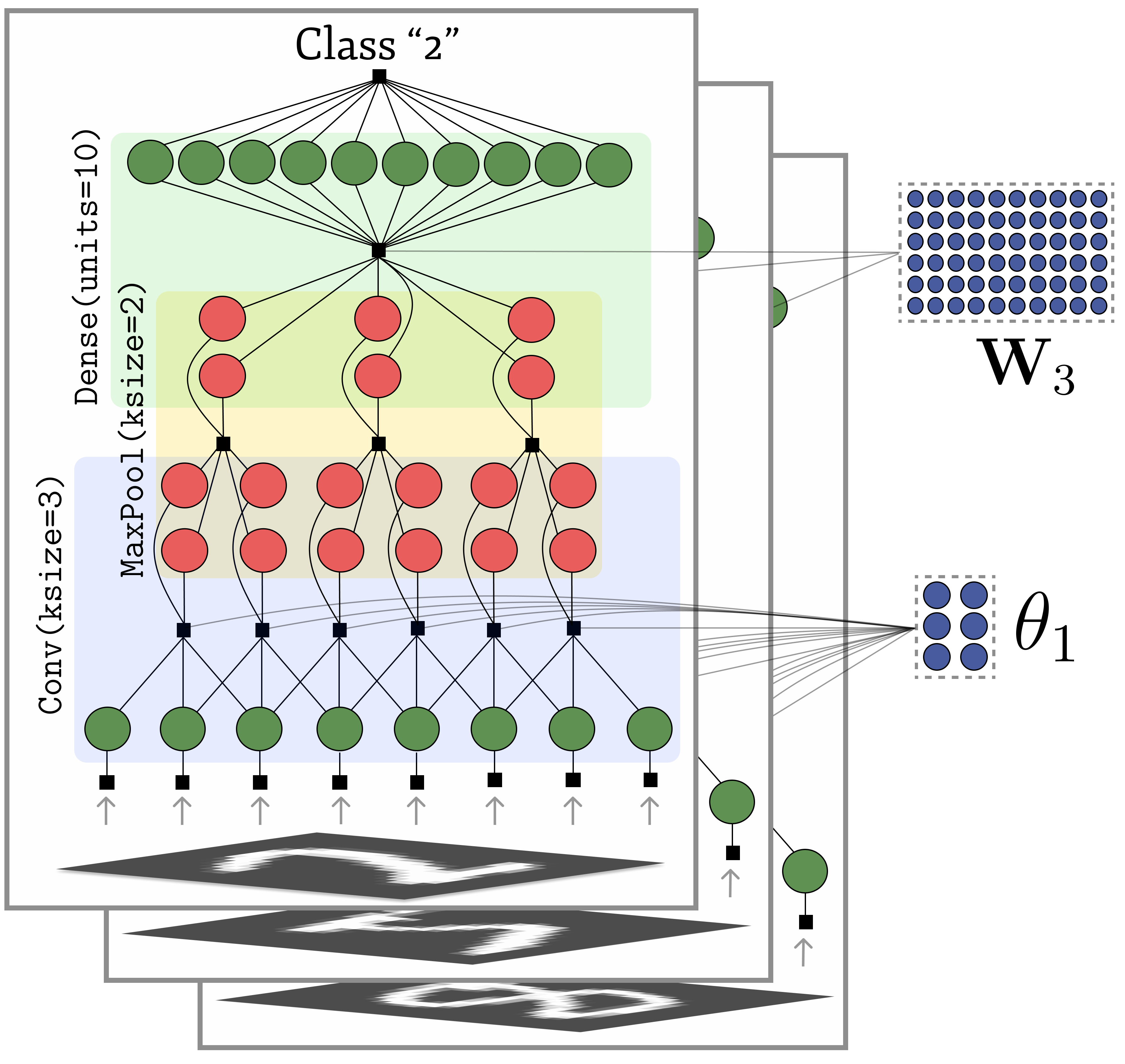}
    \caption{In {\bf \acrshort*{gbp} Learning}, we design factor graphs whose structure mirrors common NN architectures, enabling distributed training and prediction with GBP. Learnable \textcolor{filtervarcol}{parameters} are included as random variables (circles), as are \textcolor{pixelvarcol}{inputs, outputs} and \textcolor{coeffvarcol}{activations}. The \textcolor{filtervarcol}{parameters} are shared over across all observations, where the other variables are copied once per observation. Factors (black squares) between layers constrain their representations to be locally consistent, while those attached to inputs and outputs encourage compatibility with observation. The inter-layer factors are non-linear to enable soft-switching behaviour. This example architecture for image classification comprises convolutional, max pooling and dense projection layers. %
    The same architecture could be trained without supervision by removing the output observation factor.}

    \label{fig:conv_factor_graph}
    \vspace{-10pt}
\end{figure}

Deep learning (DL) has been transformative across many domains. However, its applicability is limited in cases where i) we require efficient, robust representations which can be trained incrementally; ii) supervision is sparse or irregular; and iii) learning must augment, or run alongside, hand-designed solvers. Pretrained models, when available, might mitigate these limitations, but struggle as train and test distributions diverge. How they should be updated online remains an open research question.

Concurrently, we reflect on how neural networks (NNs) are trained. Despite backpropagation~\cite{rumelhart1985learning} being largely successful for DL, training NNs spread over multiple processors is made difficult by \emph{backward locking}: processors for earlier layers sit idle after their part in the forward pass, awaiting the backward error signal. %
This challenge will become more pertinent with the growth of i) larger models which must be distributed over many devices, ii) new hardware architectures whose cores have significant local memory~\citep{Graphcore,cerebras}, and iii) parallel, distributed and heterogeneous embedded devices~\citep{Sutter:Jungle2011}. We thus expect a growing need for more flexible training algorithms which admit efficient model-parallelism.

We argue the above challenges essentially relate to the fusion of multiple signals: old and new (for incremental learning); hand-crafted vs learnt; and representations between different layers of a model (distributed training). Bayesian principles offer a clear answer on how to fuse: signals should be combined according to the rules of probability. 
We seek to exploit this fact in our probabilistic approach to DL. Specifically, our models are factor graphs (Fig.~\ref{fig:conv_factor_graph}) with random variables for all quantities relevant to DL: inputs, outputs, activations and parameters. This representation enables continual learning via online updating of the parameter posterior, and interoperability through connections with other factor graphs. We seek to design the models so as to retain the properties which we believe make DL powerful. Namely random initialisation and over-parameterisation \citep{allen2019learning}; architectural motifs encoding good inductive biases; and non-linearities which switch to selectively activate and prune when exposed to data \citep{glorot2011deep}.

Much recent work has shown that Gaussian BP (GBP) to be a robust and effective algorithm for distributed inference in factor graphs, even in the presence of non-linear and non-Gaussian factors~\citep{Davison:Ortiz:ARXIV2019,Ortiz:etal:CVPR2020,Murai:etal:ARXIV2022,Patwardhan:etal:ARXIV2022}. It is thus our choice of inference engine here. By approximating the factors in our model as Gaussian, we can use GBP for training (inference over parameters, given observations) and prediction (inference of e.g. outputs, given parameters and inputs). The generality of GBP provides flexibility as to which variables are observed, meaning training and prediction are essentially the same computation, and partial observations or missing labels do not require fundamentally different treatment.  %
As BP is inherently local and stateful, our training can be distributed and asynchronous. %

Our work shares similar goals to \citenop{lucibello2022deep}, who train MLP-like factor graphs using GBP with analytically derived message updates. For computational efficiency, they focus on architectures with binary weights and sign activation functions. In contrast, our approach enables training with GBP on arbitrary architectures, without rederivation of the message updates; and we focus on models with continuous weights in natural analogue to NNs. We demonstrate this generality, training convolutional architectures with our approach and showing we can outperform \citenop{lucibello2022deep} on image classification tasks. %

We call our approach {\bf \acrlong*{ourmethod}}. Our models (Section~\ref{sec:deep_factor_graph}) are factor graphs with architectures inspired by those in DL, and factors between layers to enforce multi-layer consistency as used in the predictive coding literature \cite{millidge2022predictive}. Within these models, GBP admits flexible and distributed training and prediction (Section~\ref{sec:method_gbp}), and learning can be done incrementally via Bayesian filtering over parameters (Section~\ref{sec:continual_learning}). Our experiments demonstrate the benefit of factor graphs with learnable parameters over hand-designed solvers in a video denoising task (Section~\ref{sec:video_denoising}). For image classification (Section~\ref{sec:img_classification}), we evaluate our continual learning approach by single epoch training on MNIST. We achieve performance equivalent to an Adam-trained CNN with a replay buffer of $6\times10^3$ examples, and show this performance is robust to asynchronous training. Last, we compare to \citenop{lucibello2022deep}, outperforming them on both MNIST (by $0.8\%$) and CIFAR10 (by $11.8\%$).

Our main contributions are as follows:
\begin{enumerate}[itemsep=0pt,parsep=5pt,topsep=0pt]
    \item An {\bf approach to train deep factor graphs with GBP}, which can be applied to any architecture and supports incremental learning.
    \item {\bf Experimental results} for convolutional architectures which show promise for continual image classification and video denoising.
\end{enumerate}

\section{Background}
\label{sec:background}
\subsection{Factor graphs}

A factor graph is a probabilistic graphical model which defines a joint distribution over variables $\allvars$ as the product of factors $\allfacs=\{\fac_j\}$:
\begin{align}
	\p\lrb{\allvars} &= \frac{1}{Z}\prod_{j=1}^{|\allfacs|} \fac_j \lrb{\varbffac{j}}
 ~.
\end{align}
Here $\varbffac{j}$ denotes the vector of all $\nconnvar_j$ variables in the neighbourhood of $\fac_j$ and $Z$ is a normalising constant. A factor graph is bipartite: variables only connect to factors and vice versa. Each factor may encode an observation, or prior on one or many variables. Factor functions $\fac_j(\cdot)$ %
may be unnormalised distributions relating to their energy, $E_j(\cdot)$, as %
\begin{align}
    \fac_j \lrb{\varbffac{j}} &= \exp\lrb{-E_j\lrb{\varbffac{j}}}.
\end{align}
A Gaussian factor graph is one in which all $\{E_j\}$ are quadratic in the related observation $\obs_j$, i.e.
\begin{align}
    E_j\lrb{\varbffac{j}}=\frac{1}{2}\lrb{\obs_j - \measfn{\varbffac{j}}}^{\tpose}{\Prec_{\obs_j}}\lrb{\obs_j - \measfn{\varbffac{j}}} \label{eq:Egauss}
\end{align}
where $\measfn{\cdot}$ is the measurement function, and $\Prec_{\obs_j}$ is the measurement precision. Note that $\obs_j$ may be a ``pseudo-observation'', e.g. the mean of a prior on $\varbffac{j}$.

\subsection{Belief Propagation}
Belief Propagation \citep{pearl1988probabilistic} is a message passing algorithm to perform inference on factor graphs via distributed, iterative computation. Each message is passed along the edge between one factor $\fac_j$ and one variable $x_i$. Messages travel in both directions, and we use the notations $\msg{\fac_j}{x_i}$ and $\msg{x_i}{\fac_j}$ for the two message types. Each message is a probability distribution in the space of the variable involved. The following update rules are iterated until convergence:
\begin{align}
    \msg{\var_i}{\fac_j}\lrb{\var_i}&\leftarrow \prod_{k\in \nei{i}\setminus j} \msg{\fac_k}{\var_i}\lrb{\var_i}\label{eq:var_to_fac_update}\\
    \msg{\fac_j}{\var_i}\lrb{\var_i}&\leftarrow \!\!\!\!\!\!\sum_{\{\var_n\}_{\nei{j}\setminus i}}\!\!\!\!\!\fac_j\lrb{\varbffac{j}}\!\!\!\!\!\prod_{n\in \nei{j}\setminus i} \!\!\!\!\!\msg{\var_n}{\fac_i}\lrb{\var_n}~, \label{eq:fac_to_var_update}
\end{align}
where $\nei{k} \setminus l$ denotes the indices for variables connected to factor $\fac_k$, except $l$. After convergence the posterior marginal of a variable is estimated by the product of its incoming messages:
\begin{align}
    \p\lrb{\var_i} &= \frac{1}{Z_i}\prod_{j\in \nei{i}}\msg{\fac_j}{\var_i}\lrb{\var_i} \label{eq:marginal_belief}
    ~,
\end{align}
where $Z_i$ is straightforward to compute if the messages belong to a known parametric family. In this work, we assume the variable nodes are univariate, but BP can be extended for multivariate posterior inference by passing vector messages between sets of variables and factors. 

BP updates \myeqref{eq:var_to_fac_update}, \myeqref{eq:fac_to_var_update}, \myeqref{eq:marginal_belief} have a number of interesting properties. First, the required computations rely only on the local state of the graph; no global context is necessary. Second, for exponential family distributions under natural parameterisation, taking the product of a set of messages is reduced to the addition of their parameters.

The above routine \myeqref{eq:var_to_fac_update}, \myeqref{eq:fac_to_var_update} is guaranteed to converge to the correct marginals in tree-structured graphs but not for graphs containing cycles. Despite a lack of guarantees, application of BP to loopy graphs has been successful in many domains, most notably for error-correcting codes \citep{gallager1962low,mackay1997near}.

\subsubsection{Gaussian Belief Propagation}
\label{sec:gbp_background}
In Gaussian factor graphs, messages $\msg{\var_i}{\fac_j}\lrb{\var_i}$ and $\msg{\fac_j}{\var_i}\lrb{\var_i}$ are normal distributions, with natural parameters being the precision (inverse covariance) matrix $\Prec$ and information vector $\info$.
The products in the variable to factor message formula \myeqref{eq:var_to_fac_update} become sums:
\begin{equation}
\begin{aligned}
	\Precmsg{\var_i}{\fac_j} &\leftarrow \sum_{k\in \nei{i}\setminus j} \Precmsg{\fac_k}{\var_i}\lrb{\var_i}
 ~;\\
	\infomsg{\var_i}{\fac_j} &\leftarrow \sum_{k\in \nei{i}\setminus j} \infomsg{\fac_k}{\var_i}\lrb{\var_i}.\label{eq:var_to_fac_gauss}
\end{aligned}
\end{equation}
These updates can be implemented efficiently by computing the belief for $\var_i$ once \myeqref{eq:marginal_belief} and then subtracting the incoming message from each factor to get the corresponding outgoing message. For a factor $\fac_j$ with precision matrix $\Prec^{\lrb{\fac_j}}\in \R^{\nconnvar_j\times \nconnvar_j}$ and information vector $\info^{\lrb{\fac_j}}\in \R^{\nconnvar_j}$, the factor to variable messages are:
\begin{equation}
\begin{aligned}
    \Precmsg{\fac_j}{\var_i} &\leftarrow \Prec^{\lrb{\fac_j}}_{i,i} -  \Prec^{\lrb{\fac_j}}_{i,\setminus i} \Sigma_{\setminus i, \setminus i}^{\lrb{\fac_j + \msgnoedge}}\Prec^{\lrb{\fac_j}}_{\setminus i, i}~;\\
	\infomsg{\fac_j}{\var_i} &\leftarrow \info^{\lrb{\fac_j}}_{i} -  \Prec^{\lrb{\fac_j}}_{i,\setminus i} \Sigma_{\setminus i, \setminus i}^{\lrb{\fac_j + \msgnoedge}}\info^{\lrb{\fac_j+m}}_{\setminus i} \label{eq:fac_to_var_gauss}
 ~,
\end{aligned}
\end{equation}
where we have used $\Sigma_{\setminus i, \setminus i}^{\lrb{\fac_j + \msgnoedge}} = \lrb{\Prec^{\lrb{\fac_j}}_{\setminus i, \setminus i} + \lrb{\allPrecmsgs_{\fac_j}}_{\setminus i, \setminus i}}^{-1}$, $\info^{\lrb{\fac_j+m}}:=\info^{\lrb{\fac_j}}+\allinfomsgs_{\fac_j}$, $\allPrecmsgs_{\fac_j}$ for the matrix of precision messages coming into $\fac_j$, and $\allinfomsgs_{\fac_j}$ for the vector of incoming information messages. Subscript $_{\setminus r}$ indicates all elements except that for variable $r$. Note that $\allPrecmsgs_{\fac_j}$ is diagonal for graphs with scalar variable nodes.

Convergence and correctness of GBP and similar algorithms has been studied extensively \citep[e.g.][]{malioutov2006walk,moallemi2009convergence,moallemi2010convergence}. As for general models, BP is not guaranteed to converge in Gaussian models with cycles. However, if it does converge in linear-Gaussian models, it is guaranteed to converge to the correct posterior means \citep{weiss1999correctness}. Further, many iterative pre-conditioning schemes have been proposed which guarantee convergence in the linear-Gaussian case \cite{johnson2009fixing,ruozzi2013message}. The guarantees do not hold for models with non-linear factors.
We refer the reader to \citenop{ortiz2021visual} for an intuitive introduction to \acrshort*{gbp}.

\subsection{Non-linear Factors}
\label{sec:nonlinear_factors}
\acrshort*{gbp} supports the use of factors which include non-linear transformations of their connected variables, and this is crucial in our proposed use here for representation learning. %
Every time a message from a non-linear factor is computed, the factor is linearised around the current variable estimates, $\allvars_{0}$, i.e. for each factor $\measfn{\varbffac{j}}\approx \measfn{\varbffaclin{j}} + \Jac_j ^{\tpose}\lrb{\varbffac{j} - \varbffaclin{j}}$
where {\small $\Jac_j:=\left. \partial \measfn{\varbffac{j}} /\partial \varbffac{j}\right|_{\varbffaclin{j}}$}.
The resulting approximated factor has a quadratic (Gaussian) energy, and the following factor precision and information can be derived
\begin{equation}
\begin{aligned}
	\Prec^{(\phi_j)} &\approx \Jac_j^{\tpose} \Prec_{\obs_j} \Jac_j~;\\
	\info^{(\phi_j)} &\approx \Jac_j^{\tpose} \Lambda_{\obs_j} \lrb{\Jac_j^{\tpose} \varbffaclin{j} + \obs_j - \mathbf{h}\lrb{\varbffaclin{j}}}~.
\end{aligned}
\label{eq:nonlinear_gauss_factor}
\end{equation}
\myeqref{eq:nonlinear_gauss_factor} can be substituted into the message update rules \myeqref{eq:fac_to_var_gauss}, enabling GBP inference in the linear-approximated model. 

To do approximate inference in the original model, one can alternate between i) setting the linearisation point $\allvars_{0}$ to be the MAP estimate of the variables given current messages, and ii) updating messages given the current linearisation point. We emphasise that only non-linear factors need to be iteratively approximated in this way, and linear-Gaussian factors remain static throughout inference.  
We refer the reader to Section 3.3 of \cite{Davison:Ortiz:ARXIV2019} for further details on GBP in non-linear models.

\section{\acrshort*{gbp} Learning}
\label{sec:method}
Our aim is to produce factor graphs which have similar architectural inductive biases to NNs, that can be overparameterised in a similar way, but can be trained with GBP. We will now describe the key factors in our model and our efficient \acrshort*{gbp} routine for training and prediction. The factor energies of form \myeqref{eq:Egauss} contain $\measfn{\cdot}$ and $\obs$ which, along with $\Prec_\obs$, are sufficient to deduce the linearised factor \myeqref{eq:nonlinear_gauss_factor}. This, in turn, defines the GBP message updates \eqref{eq:var_to_fac_gauss}, \eqref{eq:fac_to_var_gauss}. It is thus sufficient to describe our model in terms of factor energies. We emphasise that our model parameters are included as random variables in the factor graph and inferred with GBP.

\subsection{Deep Factor Graphs}
\label{sec:deep_factor_graph}
We design networks to find representations based on local consistency, i.e. the activations $\input_l\in\R^{D_l}$ in a layer $l$ should ``predict'' those in either the previous layer $\input_{l-1}\in\R^{D_{l-1}}$ or subsequent layer $\input_{l+1}\in\R^{D_{l+1}}$, via a parametric non-linear transformation $\mathbf{f}(\cdot, \filters_l)$. Applying this principle in e.g. the generative direction, together with the Gaussian assumption, suggests the following form for the factor energy:
\begin{align}
    E(\input_{l}, \input_{l-1}, \filters_l) &= \frac{\lrbtwonorm{\input_{l-1}-\mathbf{f}(\input_{l}, \filters_l)}^2}{2\sigma_l^2} \label{eq:generative_factor_energy}
    ~,
\end{align}
which is low when $\mathbf{f}(\input_l, \filters_l)$ matches the input\footnote{We use ``inputs'' to mean the activation variables within a layer which are closer to the pixels, and ``outputs'' those which are further away. However, BP is bidirectional so these quantities are not equivalent to the inputs of a function.} $\input_{l-1}$ using parameters $\filters_l$ and output $\input_l$. $\sigma_l$ is the factor strength ($\Prec_{\obs_l}=\frac{1}{\sigma_l^2}\mathbf{I}_{D_{l-1}}$). Through choice of $\mathbf{f}(\cdot,\cdot)$ we can encode different operations and inductive biases. %

CNNs, which have been successfully applied across a range of computer vision tasks, have a sparse connectivity structure which suggests efficient factor graph analogues. In our convolutional layers, a factor at spatial location $(a,b)$ connects to i) the $\filtsize_l\times\filtsize_l$ patch of the input within its receptive field, $\inputpatch_{l-1}^{(a,b)}\in\R^{\filtsize_l\times\filtsize_l\times C_{l-1}}$, ii) the corresponding activation variable for an output channel $c$, $\input_l^{(a, b, c)}\in \R^{C_l}$  and iii) the parameters: filters $\filter_{l}^{(c)}\in\R^{\filtsize_l\times\filtsize_l\times C_{l-1}}$ and bias $\biasscal_l^{(c)}$ for output channel $c$, which are shared across the layer. The energy can be written as:
\begin{align}
    \Econv^{(a,b,c)} &= \frac{\lrb{\inputscal_{l}^{(a,b,c)} - r\lrb{\inputpatch_{l-1}^{(a,b)},\filter_{l}^{(c)}, \biasscal_l^{(c)}}}^2}{2\sigma_l^2}~, \label{eq:Econv}\\
    r\lrb{\mathbf{A},\mathbf{B},s}&:=g\lrb{\flatten{A}\cdot\flatten{B}+s}~.
\end{align}
$g(\cdot)$ is an elementwise non-linear activation function, and we have removed the functional dependence of $\Econv^{(a,b,c)}$ on the connected variables for brevity. The total energy of the layer is found by summing $\Econv^{(a,b,c)}$ over spatial locations $(a,b)$ and output channels $c$.

We can similarly define a transposed convolution layer. In this case, each filter is weighted by the output variable of its corresponding channel, and the weighted sum reconstructs the inputs. Note that for a stride smaller than the kernel size, each input will belong to multiple receptive fields which are summed over to give its reconstruction. For example, for a stride of one and neglecting edge effects, these contributions can be combined according to:
\begin{align}
    \EconvT^{(a,b,c)} &= \frac{\lrb{\inputscal_{l-1}^{(a,b,c)} - r\lrb{\inputpatch_{l}^{(a,b)},\filter_{l}^{(c)} ,\biasscal_l^{(c)}}}^2}{2\sigma_l^2}\label{eq:EconvT}~.
\end{align}
In this case, the parameters have dimension $\filter_{l}^{(c)}\in\R^{\filtsize_l\times\filtsize_l\times C_{l}}$ and $\biasscal_l^{(c)}\in\R$.

For image classification, we use a dense projection layer which translates the activations in the preceding layer $\inputpatch_{L-1} \in \R^{H_{L-1} \times W_{L-1} \times C_{L-1}}$ to a vector $\input_{L}\in\R^{C_L}$, where $C_L$ is the number of classes. The dense factor energy is:
\begin{equation}
    \Edense = \frac{\lrbtwonorm{\input_{L} - g\lrb{ \weight_{L} \tpose \flatten{\inputpatch_{L-1}} + \densebias_L}}^2}{2\sigma_L^2} \label{eq:Edense}
    ~.
\end{equation}
where the activation function $g(\cdot)$ is included for generality, but usually chosen to be identity for a last-layer classifier.

In addition to convolutional and dense factors, we design factors akin to other common CNN layers. For example, we use max-pooling factors to reduce the spatial extent of the representation, upsampling layers to increase it and softmax observation factors for class supervision. We include the energies for these factors in App.~\ref{sec:app:other_factor_energies}. Note the the set of layers described here is non-exhaustive, we leave the exploration of other layer types as future work.

These ``layer'' abstractions can be composed to produce deep models with similar design freedom to DL. Further, our models may be overparameterised by introducing large numbers of learnable weights and we believe that the inclusion of non-linear activation functions $g(\cdot)$ in our factor graph can aid representation learning as in DL. In particular, we note that for non-linear factors, the factor Jacobian is a function of the current variable estimates $\Jac_j=\Jac_j\lrb{\varbffaclin{j}}$, which will cause the strength of the factors \myeqref{eq:nonlinear_gauss_factor} to vary depending on the input data. We expect this to produce a similar ``soft switching'' of connections as observed in non-linear NNs.

\subsection{Learning and Predicting with GBP Inference}
\todo{Polish this subsection}\label{sec:method_gbp}
We have described the components of deep factor graph models, in which inputs, outputs, activations and parameters are random variables. As these models include non-linear factors such as \myeqref{eq:Econv}, \myeqref{eq:EconvT} and \myeqref{eq:Edense}, we use the iterative linearisation scheme described in Section~\ref{sec:nonlinear_factors} to do approximate inference with GBP. We apply this inference engine to estimate posteriors over all latent variables. Note that there is no fundamental difference between training and prediction --- only a difference in which variables are observed. In training, we infer a posterior over parameters given observed inputs and, where supervision is available, outputs. To then make predictions on new examples, we run GBP to predict the unobserved inputs/outputs given the observed inputs/outputs and parameters.

Our message schedule proceeds as follows unless stated otherwise. For a given batch, we initialise the graph and update messages by sweeping forward and backward through the layers: first nearest the input observations, progressing to the deepest layer and back again. We repeat this for a specified number of iterations. Within each layer, we compute all factor to variable updates in parallel, and likewise for the variable to factor messages of each variable type (inputs, outputs, parameters as applicable). In addition, we experimentally show our approach works well with layerwise asynchronous message schedules (Section~\ref{sec:async_training}). While we have demonstrated these schedules work, they are likely suboptimal and we leave exploration for future work. We find that applying damping \citep{murphy2013loopy} and dropout to the factor to variable messages is sufficient for stable GBP.

\subsection{Efficient GBP}
\label{sec:fac_to_var_opt}
Efficient inference is necessary for our models to be useful in practice. 
However, the inversion of a $\lrb{\nconnvar_j-1}\times\lrb{\nconnvar_j-1}$ matrix to compute $\Sigma_{\setminus i, \setminus i}^{\lrb{\fac_j + \msgnoedge}}$ in \myeqref{eq:fac_to_var_gauss} has $O\lrb{(\nconnvar_j-1)^3}$ complexity, bottlenecking GBP. To alleviate this, we exploit the structure of the matrix being inverted. In particular, i) for factors with observation dimension $M=\dim\lrb{\obs}<\nconnvar_j$, the precision $\Prec^{(\fac_j)}$ \myeqref{eq:nonlinear_gauss_factor} is low-rank, and ii) for graphs with scalar variable nodes, $\allPrecmsgs$ is diagonal. Thus the sum $\Prec^{(\fac_j)}+\allPrecmsgs_{\fac_j}$ may be efficiently inverted via the Woodbury identity \citep{woodbury1950inverting}. Further savings come from reusing intermediates when computing messages to multiple variables \myeqref{eq:nonlinear_gauss_factor}. These optimisations change the complexity of updating messages from a factor to all $V$ variables, from $O(V\lrb{V-1}^3)$ to $O\lrb{VM^3}$. Space complexity is changed from $O(V^2)$ to $O(VM+M^2)$. See App. \ref{sec:app:fac_to_var_efficiency} for details. 

These results constitute significant savings when $M<<V$, raising the question of typical values of observation dimension $M$. For feedforward factors such as \myeqref{eq:Econv} and \myeqref{eq:Edense}, $M$ is equal to the number of output variables connected to the factor. Thus for layers with many outputs, the cubic complexity of the factor to variable update with $M$ may be prohibitive. However, we note that such factors can be decomposed along the output dimension. For example, a dense factor with energy \eqref{eq:Edense} may be decomposed into $M$ smaller factors, one per output variable. 
As the energy of the original dense factor is recovered by summing the decomposed factor energies, $\Edense=\sum_j \Edensesub{j}$, the model is remains unchanged. However, each message update is $M^2$ more efficient because one factor with $M$ outputs has been replaced by $M$ factors each with one output. 

We now consider how the factor to variable message update complexity translates to the complexity of updating \emph{all} the messages in a model. As an illustrative example, we consider a model comprising $L$ dense layers, each with $C$ input units and $C$ output units. As described, we can decompose the factor between each pair of layers into $C$ smaller factors, each with $M=1$ and connected to $2\cdot(C+1)$ variables. The complexity of factor to variable message updates in a layer is therefore $O(C^2)$, the same as the variable to factor message updates. In the non-linear case, we must also account for the computation of the factor Jacobian $\Jac$ and information vector $\info$ each time the factor is relinearised. This requires the matrix-vector product $\weight_{L} \tpose \flatten{\inputpatch_{L-1}}$, which is also complexity $O(C^2)$. We thus conclude that the overall complexity for updating all messages in a model of $L$ such layers, with a batch size of $B$, is $O(BLC^2)$. This is the same complexity as a forward or backwards pass of backpropagation in the equivalent MLP, and the same as that for the approach of \citenop{lucibello2022deep}.

\subsection{Continual Learning and Minibatching}
\label{sec:continual_learning}
We now describe how we can do continual learning of model parameters with Bayesian filtering. For generality, we use $\allparams=\{\layerparams_l\}_{l=1}^L$ to denote the set of all parameters where $\layerparams_l$ is the vector of those for layer $l$. After initialising parameter priors $p_{t=1}\lrb{\layerparams_{l,i}}\leftarrow \mathcal{N}\lrb{0,\sigma}$ we perform the following for each task $t$ in a sequence of datasets $[\inputobs_1,\ldots,\inputobs_T]$:
\begin{enumerate}[itemsep=0pt,parsep=5pt,topsep=0pt]
    \item construct a copy of the graph with task dataset $\inputobs_t$,
    \item connect a unary prior factor to each parameter variable, equal to the marginal posterior estimate from the previous task $p_t\lrb{\layerparams_{l,i}}\leftarrow p\lrb{\layerparams_{l,i}|\inputobs_{1:t-1}}$,
    \item run \acrshort*{gbp} training to get an estimate of the updated posterior $p(\allparams)=\prod_l\prod_{i=1}^{|\layerparams_l|} p\lrb{\layerparams_{l,i}|\inputobs_{1:t}}$.
\end{enumerate}
This method is equivalent to doing message passing in the combined graphical model for all tasks, but where messages between tasks are only passed forward in $t$. The advantage however, is that datapoints can be discarded after processing, and the combined graphical model for all tasks does not have to be stored in memory. As such, we also use this routine for memory-efficient training by dividing the dataset into minibatches and treating each minibatch as a task.

\section{Related Work}
\label{sec:related}
Our models can be viewed as probabilistic energy-based models \citep[EBMs;][]{lecun2006tutorial,du2019implicit} %
whose energy functions are the sum of the quadratic energies for all factors in the graph. The benefit of using a Gaussian factor graph is a model which normalises in closed form without resorting to expensive MCMC sampling. While the quadratic energy may seem constraining, we add capacity with the introduction of non-linear factors and overparameterisation. For most EBMs, parameters are attributes of the factors, where we include them as variables in our graph. As a result, learning and prediction are the same procedure in our approach, but two distinct stages for other EBMs.

Of the EBM family, restricted Boltzmann machines \citep[RBMs][]{smolensky1986information} are of particular relevance. Their factor graph resembles a single layer, fully connected version of our model, however RBMs are models over discrete variables which are unable to capture the statistics of continuous natural images. While exponential family generalisations exist, such as Gaussian-Bernoulli RBMs \citep{welling2004exponential}, %
scaling RBMs to multiple layers remains a challenging problem. To our knowledge there are no working examples of training stacked RBMs jointly, instead they are trained greedily, with each layer learning to reconstruct the activations of the trained and fixed previous layer \citep{hinton2006fast,hinton2006reducing}. As an artefact of this, later layers use capacity to learn artificial correlations induced by the early layers, rather than correlations present in the data. Global alignment is then found by fine tuning with either backprop \citep{hinton2006reducing} or wake-sleep \citep{hinton2006fast}.
We have found GBP Learning to work in multiple layer models without issue.

Our approach also relates to Bayesian DL \citep{neal2012bayesian} as we seek to infer a distribution over parameters of a deep network. Common methods to train Bayesian NNs are based on e.g. variational inference \citep{blundell2015weight, gal2016dropout}, the Laplace approximation \citep{mackay1992practical,ritter2018scalable}, Hamiltonian Monte Carlo \citep{neal2012bayesian}, Langevin-MCMC \citep{zhang2019cyclical}. Each of these  comes with benefits and drawbacks, but we note two distinguishing features of our method. Computationally, prior Bayesian DL methods rely on backprop, and so inherit its restrictions to distributed and asynchronous training, which do not limit GBP Learning. Further, our activations are random variables, allowing us to model (and resolve) disagreement between bottom-up and top-down signals. %

The factor graphs we consider relate to multi-layer predictive coding (PC) models \cite{millidge2022predictive,rao1999predictive,friston2005theory,buckley2017free, alonso2022theoretical}. Designed as a hierarchical model of biological neurons in the brain, multi-layer PC networks are trained by minimising layerwise-local, Gaussian prediction errors similar to our inter-layer consistency factors. Like ours, they also include likelihood factors which ensure observed input/output variables remain close to the observation value. Moreover, some works have noted the suitability of PC for layerwise-parallelism and emerging hardware platforms \cite{salvatori2023brain}. Our framework provides a new method to train PC models with GBP, an alternative to the standard approach using gradient descent of a variational free energy \cite{millidge2022predictive}. The work of \citenop{parr2019neuronal} is an exception, and uses BP for inference in a model of biological neurons. However, they assume the model parameters are known, where we infer parameters jointly with activations. Moreover, we consider larger scale machine learning applications. Some PC works perform \emph{active inference} \cite{buckley2017free,friston2009reinforcement}, where an agent may use a PC model to choose actions in uncertain environments. We do not consider this here, but highlight that GBP Learning could be extended for action selection in a similar manner.

Much work has focused on augmenting BP with DL \citep{nachmani2016learning,satorras2021neural,yoon2019inference,lazaro2021query}. In contrast, our method does DL \emph{within} a BP framework. \citenop{george2017generative} propose a hybrid vision system in which visual features and graph structure are learnt in a separate process to the discrete BP routine used to parse scenes (given features and graph). The work of \citenop{lazaro2016hierarchical} is more similar to ours in that parameters are treated as variables in the graphical model and updated with BP. However, their factor graph comprises only binary variables, making it ill-suited to natural images. In addition, max-product BP is used to find a MAP solution, and not a marginal posterior estimate. This precludes incremental learning via Bayesian filtering.

Most relevant to our work is that of \citenop{lucibello2022deep}, who use GBP to train MLPs. Their method relies on analytically derived message updates which only apply to dense architectures, and they focus on models with binary weights and sign activations. In contrast, our approach is straightforward to apply to any network structure without rederivation of messages, assuming the appropriate factor energies can be specified. Further, we focus on continuous weights in direct analogy to NN parameters. Though they similarly minibatch by filtering over parameters, they visit each datapoint multiple times and run a small number of GBP iterations during each visit. After so few iterations, the resulting messages are unlikely to constitute an accurate posterior, and ad-hoc ``forgetting'' factors are necessary to avoid overcounting data seen multiple times. In contrast, we train with only a single epoch, running GBP to convergence on each batch, enabling straightforward filtering.

\section{Results}
\label{sec:results}
We demonstrate our approach with three experiments: some small regression tasks, sequential video denoising and image classification. Our TensorFlow \citep{Tensorflow:MISC2015} implementation is made available\footnote{\href{https://github.com/sethnabarro/gbp_learning/}{github.com/sethnabarro/gbp\_learning/}}.%

\begin{figure}[t]
    \centering
    \begin{subfigure}{0.19\textwidth}
        \centering
        \includegraphics[width=\textwidth]{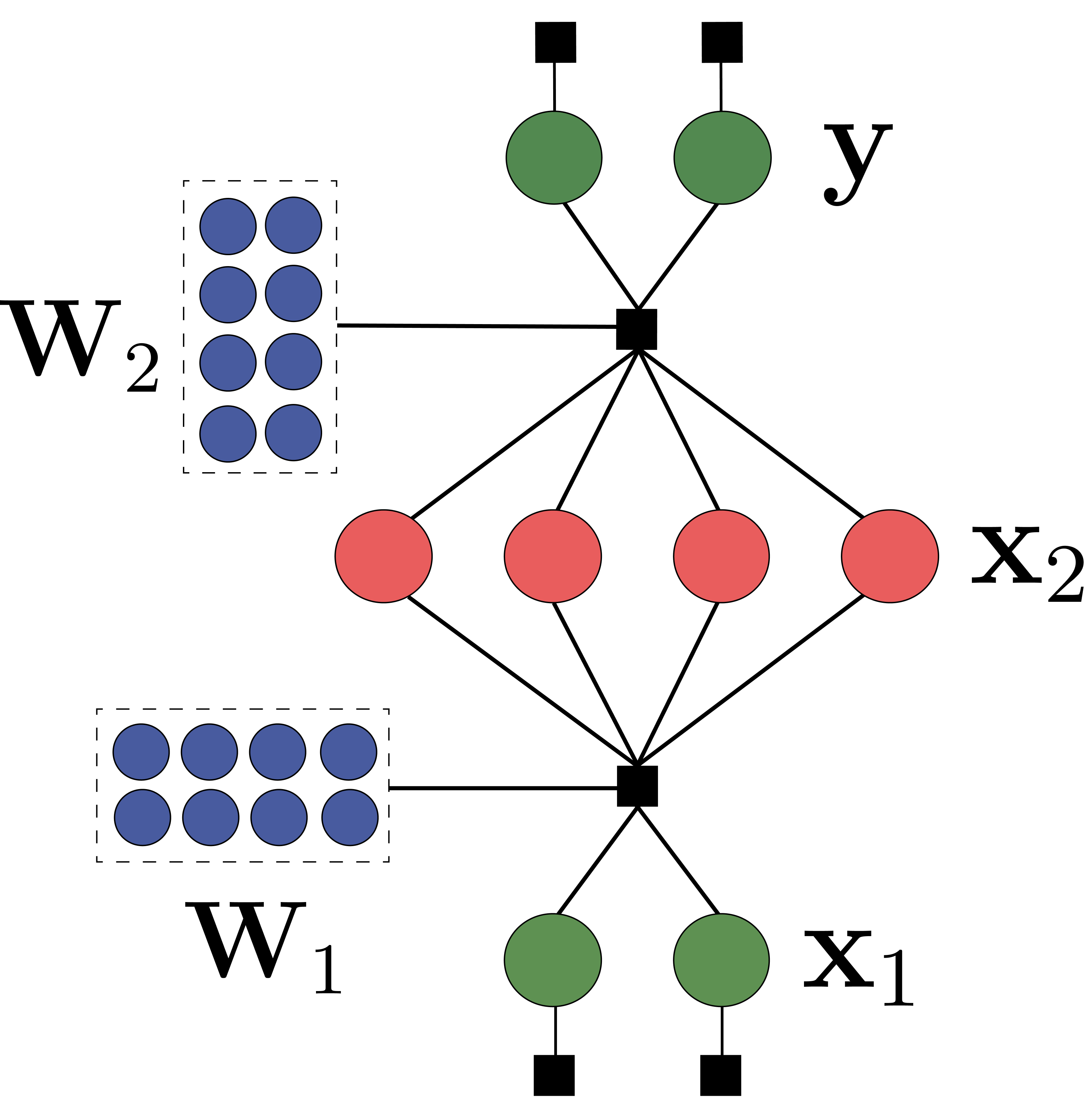}
        \vspace*{2pt}
        \caption{Architecture}
        \label{subfig:mlp}
    \end{subfigure}\hfill
    \begin{subfigure}{0.25\textwidth}
        \centering
        \includegraphics[width=\textwidth]{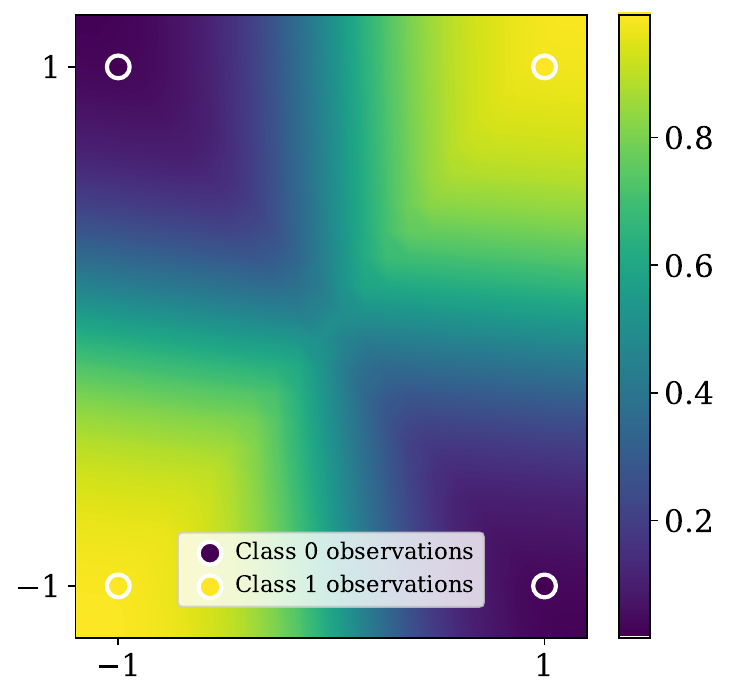}
        \caption{XOR, $p(y=1|\mathbf{x}_1,\mathbf{W}_1,\mathbf{W}_2)$}
        \label{subfig:xor}
    \end{subfigure}\\\vspace{5pt}
    \begin{subfigure}{0.43\textwidth}
        \centering
        \includegraphics[width=\textwidth]{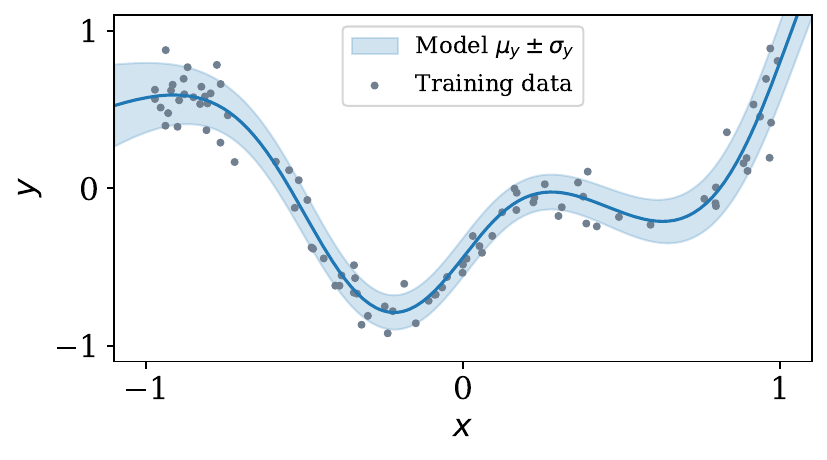}
        \caption{Nonlinear regression}
        \label{subfig:nonlin_regression}
    \end{subfigure}
    \caption{GBP Learning in MLP-like factor graphs (\subref{subfig:mlp}) can solve nonlinear regression and classification tasks. (\subref{subfig:xor}) was generated with $8$ hidden units, (\subref{subfig:nonlin_regression}) with $16$ hidden units.}
    \label{fig:small_experiments}
    \vspace{-10pt}
\end{figure}
\subsection{Toy Experiments}
We start by verifying that our model can solve tasks requiring nonlinear modelling. To this end, we run GBP Learning with single hidden layer, MLP-like factor graphs (Fig.~\ref{subfig:mlp}) in two settings. The first is ``Exclusive-OR'', which is a well-known minimal test for nonlinear modelling (Fig.~\ref{subfig:xor}). The second is a nonlinear regression problem (Fig.~\ref{subfig:nonlin_regression}) with $90$ training points. In both cases, we employ a nonlinear activation function in the first dense layer. Full details are provided in App.~\ref{sec:app:toy_experiment_details}. It is clear that GBP Learning can solve both tasks, confirming  that the linearisation method described in Section~\ref{sec:nonlinear_factors} is sufficient to capture nonlinear dependencies. 

\subsection{Video Denoising}
We now ask whether the learnable components in our model can improve performance over a hand-designed solver. We apply our method to the task of denoising the ``bear'' video from the DAVIS dataset\footnote{Creative Commons Attributions 4.0 License} \citep{perazzi2016benchmark}, downsampled to $258\times454$ with bilinear interpolation. We sample $10\%$ of the pixels in each frame of the video, replacing their intensities with noise drawn from $U(0, 1)$. Performance is assessed by how well the denoised image matches the ground truth under the peak signal-to-noise ratio (PSNR). %

The estimated pixel intensities may explained by the noisy pixel observations and/or the model reconstruction. Both the pixel observation and reconstruction factors have robust energies (see Section 5 of \citenop{Davison:Ortiz:ARXIV2019}) to enable the pixel variables to ``switch'' between these explanations. 

We evaluate two types of reconstruction model: i) Gaussian factor graphs with learnable parameters, trained with GBP Learning; and ii) a hand-designed baseline, with no learnable parameters. In the learnable models, we examine the impact of model depth by comparing: i) a single transposed convolution layer (factor energies as per \myeqref{eq:EconvT}) with four filters, and ii) a five layer model comprising transposed convolution and upsampling layers (for details see App.~\ref{sec:app:video_denoising_experiment}). In the baseline model, neighbouring pixels are encouraged to have similar intensities via shared smoothness factors (see e.g \citenop{ortiz2021visual}). We refer to this as ``pairwise smoothing''. Robust energies on the smoothness factors aid the preservation of edges present in the original image. %

In both the pairwise smoother and GBP Learning, we do inference with \acrshort*{gbp}. Note that in our models, GBP is jointly estimating the true pixel intensities \emph{and} the model parameters. We emphasise that our model learns parameters \emph{from the noisy images only} and without supervision. It is incentivised to do so by the reconstruction factors in noiseless regions. To evaluate our continual learning approach, we try two routines to infer parameters: i) learn them from scratch on each frame and ii) learn them incrementally, by doing filtering on the parameters as described in Section~\ref{sec:continual_learning}. 

The hyperparameters for all models were tuned using the first five frames of the video. Further details of the final models can be found in App.~\ref{sec:app:video_denoising_experiment}. Denoising the entire $82$-frame video with the single layer model took $\sim8$mins on a NVIDIA RTX 3090 GPU, and the five layer model took $\sim27$mins. Pairwise smoothing took $\sim2$mins.

\begin{figure}[t]
    \centering
    \includegraphics[width=0.45\textwidth]{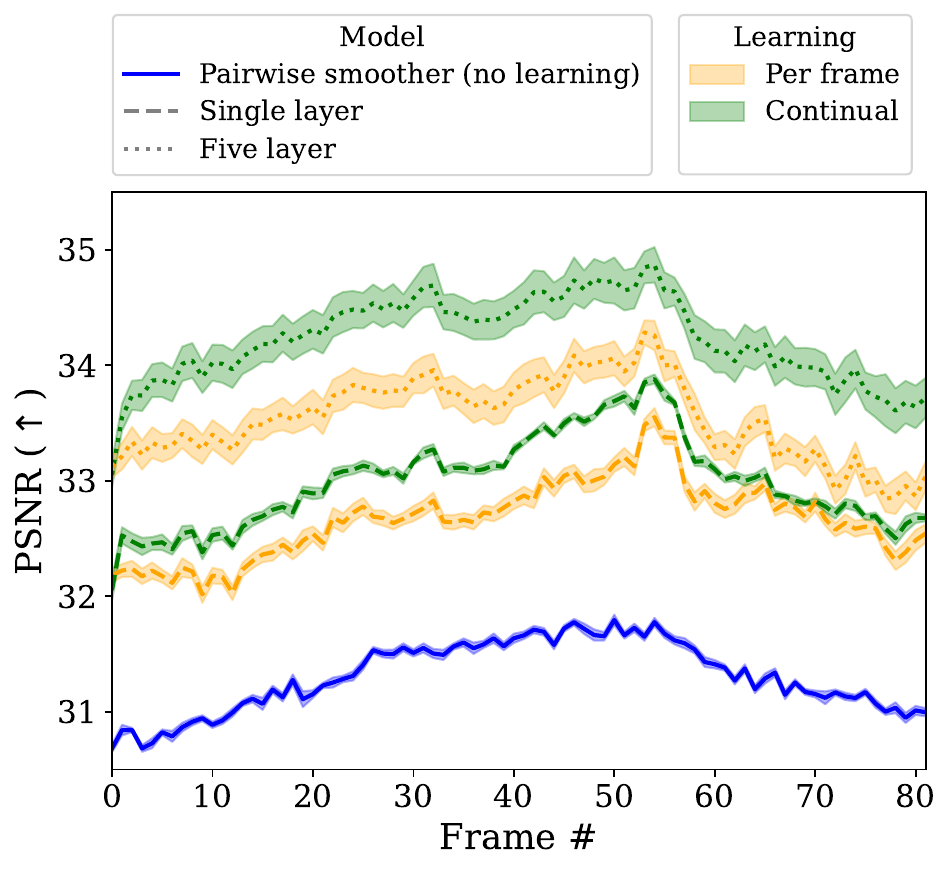}
    \vspace{-5pt}
    \caption{{\bf Video denoising results.} Factor graphs with learnable components outperform a hand-specified pairwise smoother. Continual learning of parameters over the video further improves the PSNR over per-frame learning, and the deep model outperforms the single layer. Shading is $\pm1$ standard error (SE) over $10$ seeds.}
    \label{fig:video_denoising}
    \vspace{-10pt}
\end{figure}

The PSNR results are presented in Fig.~\ref{fig:video_denoising} (and examples of denoised frames in Fig.~\ref{fig:video_denoising_frames_crop}). All methods exhibit a similar relative evolution of PSNR over the course of the video, slowly rising until the fiftieth frame and then falling again. We conjecture that this is due to varying image content, with some frames being easier to denoise than others. 

The models with learnable parameters (orange and green) significantly outperform the classical baseline (blue). We attribute this to the learned models being able to capture image structure, which enables more high-frequency features to be retained while removing the corruption.
\label{sec:video_denoising}

We see a clear benefit of depth, with significantly higher PSNR scores for the five layer model (orange/green dotted) over the single layer (orange/green dash). Despite the single layer model having only four filters, we find that it sometimes learns to reconstruct noise when learning continually over the video. To counteract this, we set the prior for each frame to be an interpolation between the previous posterior and the original prior (plotted; see App.~\ref{sec:app:video_denoising_experiment}). In contrast, the deeper model does not require additional regularisation when trained continually. These results imply i) deep factor graphs have better inductive biases, capturing higher-level patterns in the data; ii) GBP Learning is can find aligned multi-layer representations with only local message updates. 

In general we see that continual learning (green) leads to an improvement over learning per-frame (orange), suggesting that we can effectively fuse what has been learnt in previous frames and use it to better denoise the current frame.
\subsection{Image Classification}
\label{sec:img_classification}
Next, we assess our method in a supervised learning context, evaluating it on MNIST\footnote{\href{http://yann.lecun.com/exdb/mnist/}{yann.lecun.com/exdb/mnist/}. Creative Commons Attribution-Share Alike 3.0 license.}. The goal here is to gauge both sample efficiency and the ability to learn continually with only one pass through the training set. 

We train a convolutional factor graph with architecture similar to Fig.~\ref{fig:conv_factor_graph}: a convolutional layer (with energy as per \myeqref{eq:Econv}), followed by a max-pool \myeqref{eq:Emaxpool}, dense layer \myeqref{eq:Edense} and (at train time) a class observation factor \myeqref{eq:Esoftmax} on the output logit variables. Training is minibatched via the Bayesian filtering approach of Section~\ref{sec:continual_learning}, so the model only sees each datapoint once after which it can be discarded. To assess sample efficiency, we train on subsets of varying sizes, as well as the full training set. We compare against two baselines:
\begin{enumerate}[itemsep=0pt,parsep=5pt,topsep=0pt]
    \item a linear classifier baseline which predicts logits via a dense projection of the image vector, trained with Adam over multiple epochs, and
    \item the CNN equivalent of our factor graph, trained with Adam~\citep{kingma2014adam} for a single epoch, both i) with a FIFO replay buffer (varying sizes) and ii) without replay.
\end{enumerate} 
The first baseline allows us to verify our method can do nonlinear learning on images. Further, varying the size of the replay buffer in the second baseline gives an indication of how much information our model can fuse via filtering.
All model hyperparameters, including factor strengths $\sigma$, were tuned on validation sets generated by randomly subsampling $15\%$ of the training set. Further details of the models and hyperparameter selection are included in App.~\ref{sec:app:mnist_experiment}.

\begin{figure}[t]
    \centering
    \includegraphics[width=0.48\textwidth]{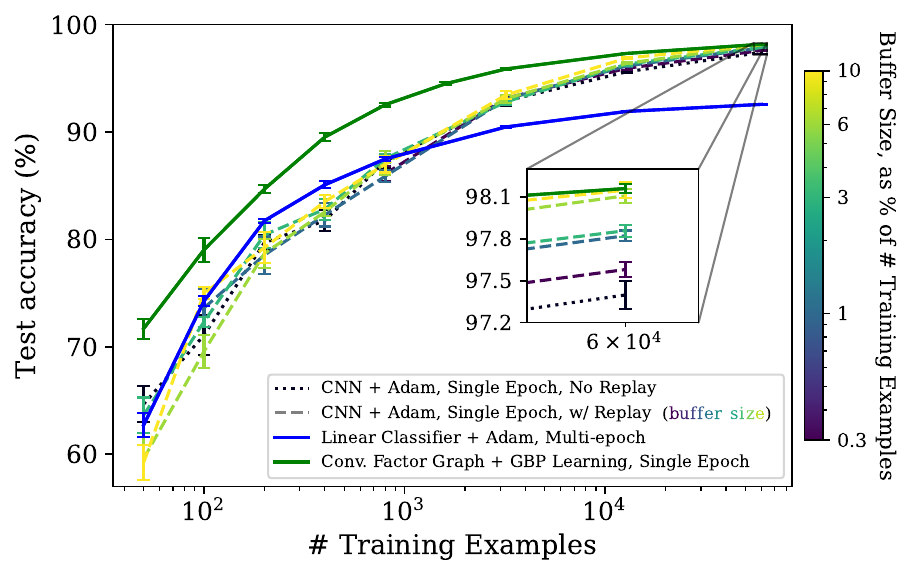}
    \vspace{-10pt}
    \caption{{\bf Single epoch MNIST results.} GBP Learning outperforms other methods in the small data regime, and performs similarly to a CNN with a replay buffer of $6\times10^3$ examples on the full training set. Error bars cover $\pm1$SE over $5$ seeds.}
    \label{fig:mnist}
    \vspace{-10pt}
\end{figure}

In the low data regime, GBP Learning comprehensively outperforms all baselines (Fig.~\ref{fig:mnist}), likely due to regularising priors and marginalising over uncertain activations. With the full training set, GBP Learning achieves a test accuracy of $98.16\pm0.03\%$, significantly higher than the linear classifier. Further, we outperform most CNN + Adam variants except those with large replay buffers ($3600$ examples/$6\%$ of training data, $6000$ examples/$10\%$ training data), to which we perform similarly. We are thus encouraged that our model can learn complex relationships incrementally, consolidating new examples into its parameter posterior online. 

Full dataset training with GBP Learning takes $\sim3$ hours on NVIDIA RTX3090 GPU. While this is much slower than the few minutes to train the CNNs on CPU, we note that the software to train NNs with backprop has benefited from decades of optimisation, as have well-established GPU hardware platforms. In contrast, our current GBP Learning implementation runs on a hardware and software stack optimised for DL. We believe orders of magnitude efficiency could be gained by developing on a more tailored setup with optimised software and hardware with on-chip memory. See Section~\ref{sec:conclusion} for discussion.

\subsubsection{Asynchronous Training}
\label{sec:async_training}
To demonstrate the robustness to different update schedules,
we also trained in a layer-wise asynchronous manner. At each iteration, we uniformly sample a sequence of $L=4$ layers with replacement to determine the layer update order. Sampling with replacement means some layers may not be updated in an iteration, where some may be updated multiple times. Such an approach yields a final accuracy of $98.11\pm0.04\%$ ($\pm1$SE over $5$ seeds), close to $98.16\pm0.03\%$ achieved by the same model trained with synchronous forward/backward sweeps (details and plot of convergence in App.~\ref{sec:app:mnist_async}). This result implies that GBP Learning can work well when the model is distributed over multiple processors, without the need for global synchronisation.
\subsubsection{Comparison with \citenop{lucibello2022deep}}
To understand how our method compares to previous work, we evaluate on the same image classification benchmarks presented in \citenop{lucibello2022deep}: MNIST, FashionMNIST and CIFAR10. \citenop{lucibello2022deep} only support dense models where our method works with general architectures. We thus compare the convolutional model described in Section~\ref{sec:img_classification} with the dense factor graph results from \citenop{lucibello2022deep}. For GBP Learning, we retain the same architecture and hyperparameters used for MNIST — we do not conduct any task-specific tuning. 

The results are given in Table~\ref{tab:img_class_lucibello}. The approaches perform similarly for FashionMNIST, but our method achieves higher accuracy for both MNIST and CIFAR10. In the latter case, we outperform \citenop{lucibello2022deep} by more than $10\%$. Moreover, we obtain such performance with incremental training, passing over the training set once only, where \citenop{lucibello2022deep} train for $100$ epochs. 

We note a significant difference in the relative improvement between FashionMNIST ($0\%$) and CIFAR10 ($11.8\%$), which we conjecture may result from CIFAR10 benefiting more from translation equivariance of the convolutional layer. In FashionMNIST, the objects of interest are mostly centred and fill the frame, where in CIFAR10 the objects are smaller and occur in different regions of the images. 
\begin{table}[t]
    \centering \small
    \caption{Image classification test accuracy ($\%$). Ranges cover $1$SE either side of the mean for $5$ seeds.}
    \label{tab:img_class_lucibello}
    \begin{tabular}{lcc}
        \toprule
        Dataset  & \citenop{lucibello2022deep} & Ours \\\midrule
         MNIST & $97.40\pm0.04$& $\mathbf{98.16\pm0.03}$\\
         FashionMNIST& $88.2\pm0.1$& $88.2\pm0.1$\\
         CIFAR10 &$41.3\pm0.1$ & $\mathbf{53.1\pm0.3}$\\\bottomrule
    \end{tabular}
    \vspace{-10pt}
\end{table}

\section{Conclusion}
\label{sec:conclusion}
We have introduced \acrshort*{gbp} Learning, a method for learning in Gaussian factor graphs. Parameters are included as variables in the graph and learnt using the same BP inference procedure used to estimate all other latent variables. Inter-layer factors encourage representations to be locally consistent, and multiple layers can be stacked in order to learn richer abstractions. Experimentally, we have shown a shallow, learnable model improves over a hand-crafted method on a video denoising task, with further performance gains coming from stacking multiple layers. We have also trained convolutional factor graphs for image classification with GBP Learning. On MNIST, we demonstrate encouraging sample efficiency, and reach comparable single epoch performance to a CNN with a replay buffer of $6\times10^3$ examples. GBP Learning outperforms \citenop{lucibello2022deep} by $11.8\%$ on CIFAR10 and $0.8\%$ on MNIST.

While we find these initial results encouraging, we highlight scaling up GBP Learning to bigger, more complex models and datasets as an exciting future direction. Our current implementation is built on software and run on hardware heavily optimised for DL. Scaling up GBP Learning will require a bespoke hardware/software system, which can better leverage the distributed nature of GBP inference. In particular, we believe processors with memory local to the compute cores \citep{Graphcore,cerebras} to be a promising platform for BP. 
Different sections of the factor graphs could be mapped to different cores. Local message updates, between factors and variables on the same core, would be cheap and could be updated at high frequency, with occasional inter-core communication to ensure alignment between different parts of the model. Low-level GBP primitives, written in e.g. Poplar\footnote{\href{https://www.graphcore.ai/products/poplar}{graphcore.ai/products/poplar}}, could be used to ensure the message updates make best use of high-bandwidth, local memory. A similar system was applied to bundle adjustment problems with GBP \citep{Ortiz:etal:CVPR2020}, and was found to be around $24\times$ faster on IPU \citep{Graphcore} than a state-of-the-art CPU solver.

Further, this work has only considered GBP for marginal inference of scalar variables. We note that this is restrictive and prevents capturing the rich correlation structure if the energy landscape of deep networks. By including multidimensional variables in the factor graph, we would expect to mitigate this, and possibly enable more stable BP due to a reduced number of loops in the model. As higher dimensional variables incur greater computational cost per iteration, only variables which are likely to be highly correlated, e.g. the activations or weights within a layer, should be combined. 

\section*{Acknowledgements}
We are grateful to many researchers for helpful discussion relating to this work. In particular, we thank the members of the Dyson Robotics Lab, especially Riku Murai, Joseph Ortiz, Ignacio Alzugaray, Talfan Evans; members of MvdW's research group, particularly Anish Dhir; and the Vicarious team at Google DeepMind. 

SN and AJD are funded by EPSRC Prosperity Partnerships (EP/S036636/1) and Dyson Techonology Ltd.

\section*{Impact Statement}
Computationally, our method learns via local update rules, which could enable training to be parallelised over arrays of low-power devices in place of GPUs which have high power consumption. In addition, our flexibility to different message schedules could further boost efficiency by executing intra-processor updates at high frequency and inter-processor updates less often, thus reducing communication.

\bibliographystyle{icml2024}
\bibliography{refs,robotvision}

\newpage
\appendix

\onecolumn
\section{Energy Functions for Additional Factors}
\label{sec:app:other_factor_energies}
In Section~\ref{sec:deep_factor_graph}, we described the energies for factors connecting convolutional and transposed convolutional layers. Here, we provide the energies for additional factors we use in our models. 

For an input patch from channel $c$, $\inputpatch_{l-1}^{(a,b,c)}\in\R^{\filtsize\times\filtsize}$, centred at $(a,b)$, a connected max-pooling factor has energy:
\begin{align}
    \Emaxpool^{(a,b,c)} &= \frac{1}{2\sigma_l^2}\lrb{\max\lrb{\inputpatch_{l-1}^{(a,b,c)}} -\inputscal_{l}^{(a,b,c)}}^2~.\label{eq:Emaxpool}
\end{align}
To increase the spatial extent of a representation from layer $l$ to its input $l-1$, we use upsampling layers. These factors connect a single activation variable in $l$, $\inputscal_{l}^{(a,b,c)}$ to the patch within its receptive field in $l-1$, $\inputpatch_{l-1}^{(a,b,c)}\in\R^{\filtsize\times\filtsize}$. Their energy is
\begin{align}
    \Eupsample^{(a,b,c)} &= \frac{1}{2\sigma_l^2}\sum_{i,j=-\floor{\filtsize/2}}^{\floor{\filtsize/2}}\lrb{\inputscal_{l-1}^{(a-i,b-j,c)} -\inputscal_{l}^{(a,b,c)}}^2~.\label{eq:Eupsample}
\end{align}
Last, class supervision may then be incorporated by treating last layer activations $\input_L$ as logits and connecting them to a observation factor with energy
\begin{align}
    \Esm &= \frac{\lrbtwonorm{\Softmax{}{\input_L} - \mathbbm{1}_y}^2}{2\sigma_{\mathrm{softmax}}^2}
    ~, \label{eq:Esoftmax}
\end{align}
where $\mathbbm{1}_y$ is a one-hot encoding of the observed class $y$.

\section{Factor to Variable Message Update Optimisation}
\label{sec:app:fac_to_var_efficiency}
We aim to reduce the complexity of message update computations \myeqref{eq:fac_to_var_gauss}. Nai\"ve inversion of the sum of factor and message precisions $\Sigma_{\setminus i, \setminus i}^{\lrb{\fac_j + \msgnoedge}} = \lrb{\Prec^{\lrb{\fac_j}}_{\setminus i, \setminus i} + \lrb{\allPrecmsgs_{\fac_j}}_{\setminus i, \setminus i}}^{-1}$ has complexity $O\lrb{\lrb{V_j-1}^3}$. This can be reduced to $O\lrb{\lrb{V-1}M^2 + M^{3}}$ ($M:=\dim \obs$) by substituting $\Prec^{(\phi_j)}$ for the linearised factor precision \myeqref{eq:nonlinear_gauss_factor} and applying the Woodbury identity \citep{woodbury1950inverting}:
\begin{equation}
    \Covfacplusmsg_{\setminus i} = \allPrecmsgs_{\setminus i, \setminus i}^{-1} - \allPrecmsgs_{\setminus i, \setminus i}^{-1} \lrb{\Jac_{:,\setminus i}}^{\tpose} \lrb{\Prec_{\obs}^{-1} - \Jac_{:,\setminus i} \allPrecmsgs_{\setminus i, \setminus i}^{-1} \lrb{\Jac_{:,\setminus i}}^{\tpose}} ^{-1} \Jac_{:,\setminus i} \allPrecmsgs_{\setminus i, \setminus i}^{-1}\label{eq:fac_plus_msg_prec_inv}
~,
\end{equation}
where we have dropped subscripts and superscripts relating to $\phi_j$ for brevity.

Further efficiencies come from substituting \myeqref{eq:fac_plus_msg_prec_inv} back into the factor to variable message update \myeqref{eq:fac_to_var_gauss}, 
\begin{align}
    \Precmsg{\fac}{\var_i} &\leftarrow \Prec_{i,i} -  \Prec_{i,\setminus i} \lrb{\allPrecmsgs_{\setminus i, \setminus i}^{-1} - \allPrecmsgs_{\setminus i, \setminus i}^{-1} \lrb{\Jac_{:,\setminus i}}^{\tpose} \lrb{\Prec_{\obs}^{-1} - \Jac_{:,\setminus i} \allPrecmsgs_{\setminus i, \setminus i}^{-1} \lrb{\Jac_{:,\setminus i}}^{\tpose}} ^{-1} \Jac_{:,\setminus i} \allPrecmsgs_{\setminus i, \setminus i}^{-1}}\Prec_{\setminus i, i} \\
    \infomsg{\fac}{\var_i} &\leftarrow \info_i -  \Prec_{i,\setminus i} \lrb{\allPrecmsgs_{\setminus i, \setminus i}^{-1} - \allPrecmsgs_{\setminus i, \setminus i}^{-1} \lrb{\Jac_{:,\setminus i}}^{\tpose} \lrb{\Prec_{\obs}^{-1} - \Jac_{:,\setminus i} \allPrecmsgs_{\setminus i, \setminus i}^{-1} \lrb{\Jac_{:,\setminus i}}^{\tpose}} ^{-1} \Jac_{:,\setminus i} \allPrecmsgs_{\setminus i, \setminus i}^{-1}}\info_{\setminus i}^{(\fac + m)}
    ~,
\intertext{
and noting that $\Prec_{i,\setminus i}=\lrb{\Jac_{:,i}}^{\tpose}\Prec_{\obs} \Jac_{:,\setminus i}$. We can then write:}
    \Precmsg{\fac}{\var_i} &\leftarrow \Prec_{i,i} -  \lrb{\Jac_{:,i}}^{\tpose}\Prec_{\obs} \Jac_{:,\setminus i} \lrb{\allPrecmsgs_{\setminus i, \setminus i}^{-1} - \allPrecmsgs_{\setminus i, \setminus i}^{-1} \lrb{\Jac_{:,\setminus i}}^{\tpose} \lrb{\Prec_{\obs}^{-1} - \Jac_{:,\setminus i} \allPrecmsgs_{\setminus i, \setminus i}^{-1} \lrb{\Jac_{:,\setminus i}}^{\tpose}} ^{-1} \Jac_{:,\setminus i} \allPrecmsgs_{\setminus i, \setminus i}^{-1}}\lrb{\Jac_{:,\setminus i}}^{\tpose}\Prec_{\obs} \Jac_{:, i} \\
    \infomsg{\fac}{\var_i} &\leftarrow \info_i -  \lrb{\Jac_{:,i}}^{\tpose}\Prec_{\obs} \Jac_{:,\setminus i} \lrb{\allPrecmsgs_{\setminus i, \setminus i}^{-1} - \allPrecmsgs_{\setminus i, \setminus i}^{-1} \lrb{\Jac_{:,\setminus i}}^{\tpose} \lrb{\Prec_{\obs}^{-1} - \Jac_{:,\setminus i} \allPrecmsgs_{\setminus i, \setminus i}^{-1} \lrb{\Jac_{:,\setminus i}}^{\tpose}} ^{-1} \Jac_{:,\setminus i} \allPrecmsgs_{\setminus i, \setminus i}^{-1}}\info_{\setminus i}^{(\fac + m)}~.
\intertext{Right-multiplying the $\Jac_{:,\setminus i}$ before the parentheses, and left multiplying the $(\Jac_{:,\setminus i})^{\tpose}$ after gives}
    \Precmsg{\fac}{\var_i} &\leftarrow \Prec_{i,i} -  \lrb{\Jac_{:,i}}^{\tpose}\Prec_{\obs} \lrb{\mathbf{U}_{i} - \mathbf{U}_{ i} \lrb{\Prec_{\obs}^{-1} - \mathbf{U}_{i}} ^{-1} \mathbf{U}_{i} }\Prec_{\obs} \Jac_{:, i}~,
\intertext{where we have defined $\mathbf{U}_{i}:=\Jac_{:,\setminus i}\allPrecmsgs_{\setminus i, \setminus i}^{-1}\lrb{\Jac_{:,\setminus i}}^{\tpose}$. For the information update, we instead right-multiply $\info_{\setminus i}^{(\fac + m)}$}
    \infomsg{\fac}{\var_i} &\leftarrow \info_i -  \lrb{\Jac_{:,i}}^{\tpose}\Prec_{\obs} \lrb{\mathbf{T}_{i} - \mathbf{U}_{i} \lrb{\Prec_{\obs}^{-1} - \mathbf{U}_{i}} ^{-1} \mathbf{T}_{i} }~.
\end{align}
where $\mathbf{T}_{i}:=\Jac_{:,\setminus i}\allPrecmsgs_{\setminus i, \setminus i}^{-1}\info_{\setminus i}^{(\fac + m)}$. Given $\mathbf{U}_i$ and $\mathbf{T}_i$, both message updates are $O\lrb{M^3}$, i.e. independent of $V$. However direct computation of $\mathbf{U}_i$ or $\mathbf{T}_i$ has complexity $O\lrb{(V-1)M^2}$ for each outgoing variable $i$ ($\allPrecmsgs^{-1}$ is diagonal),  so the overall complexity is quadratic in $V$. We achieve linear complexity by exploiting $\mathbf{U}_i=\Jac \allPrecmsgs^{-1} \Jac^{\tpose} - \Jac_{:,i} \allPrecmsgs_{i, i}^{-1} \lrb{\Jac_{:,i}}^{\tpose}$
where $\Jac \allPrecmsgs^{-1} \Jac^{\tpose}$ can be computed once for all connected variables. Similarly, we use $\mathbf{T}_i=\Jac \allPrecmsgs^{-1} \info^{(\fac + m)} - \Jac_{:,i} \allPrecmsgs_{i, i}^{-1}\info_{i}^{(\fac + m)}$ for the information update. This reduces the complexity for updating all outgoing messages from a factor from $O\lrb{V\lrb{\lrb{V-1}M^2 + M^3}}$ to $O\lrb{VM^3}$. In addition, these optimisations require memory $O\lrb{VM + M^2}$ which, in most cases, is a saving relative to $O\lrb{V^2}$ needed to store the full factor precision.

\section{Toy Experiment Details}
\label{sec:app:toy_experiment_details}
\subsection{XOR}
We use the MLP-inspired factor graph architecture illustrated in Fig.~\ref{subfig:mlp}, with $8$ units in the hidden layer and a Leaky ReLU activation in the first-layer dense factor. The full model is described in Table~\ref{tab:xor_model}.

We run GBP in the training graph (with $4$ input/output observations) for $600$ iterations. We then fix the parameters, remove the softmax class observation layer and run GBP for $300$ iterations on a grid of $20\times20$ test input points. The last layer activation variables are then treated as the logits for class predictions. At both train and test time, we use a damping factor of $0.7$ applied to the factor to variable message updates from the dense factors.

\begin{table}[h!]
\centering
\begin{tabular}{lccc}\toprule
Layer \#                     & $1$                     & $2$ & $3$\\ \midrule
Layer type                   & Dense & Dense & Softmax               \\
Input dim.            & $2$  & $8$ & $2 $                   \\
Output dim.            & $8$  & $2$ & $2 $                   \\
Inc. bias                    & \checkmark & \checkmark & - \\
Weight prior $\sigma$        & $3.0 $ &$3.0$&-                 \\
Activation prior $\sigma$   & $5.0$   &$2.0$ & -                \\
Input obs. $\sigma$          & $0.02$ & - &-                  \\
Class obs. $\sigma$ & - &- & $0.1 $              \\
Dense recon. $\sigma$              & $0.1$ & $0.1$ & -                  \\
Activation function $g(\cdot)$  & Leaky ReLU & Linear & -   \\ \bottomrule               
\end{tabular}
\caption{The MLP-like factor graph used for the XOR experiment.}
\label{tab:xor_model}
\end{table}

\subsection{Regression}
To generate the 1D regression data, we sample the $90$ input points uniformly in the interval $(-1.0, 1.0)$. For each input point $x_i$, the output $y_i$ is generated according to
\begin{align}
    z_i &= 5\cdot\sin\lrb{6.7 \cdot x_i}+\lrb{10\cdot x_i}^2\cdot0.15+\epsilon_i~,\\
    y_i &= \frac{z_i - \mathrm{mean}\lrb{z}}{2\cdot\mathrm{std}\lrb{z}}
\intertext{where}
    \epsilon_i&\sim \mathcal{N}(0., 1.5 + x_i ^ 2)
\end{align}
and $\mathrm{mean}\lrb{\cdot}$, $\mathrm{std}\lrb{\cdot}$ are empirical estimates over the $90$ training points.

To fit the data, we use a similar MLP-inspired factor graph architecture to that illustrated in Fig.~\ref{subfig:mlp}. However, this architecture has only $1$ input variable and $1$ output variable. We use $16$ units in the hidden layer and a sigmoid activation in the first-layer dense factor. The full model is described in Table~\ref{tab:regression_model}.

We run GBP in the training graph (with $90$ input/output observations) for $4000$ iterations. We then fix the parameters, remove the output observation layer and run GBP for $1000$ iterations on $225$ uniformly spaced test input points. The last layer activation variables are then treated as the predictions for these query points. At both train and test time we use a damping factor of $0.8$ and a dropout of $0.6$ in the factor to variable message updates from the dense factors.

\begin{table}[h!]
\centering
\begin{tabular}{lccc}
\toprule
Layer \#                     & $1$                     & $2$ & $3$\\ \midrule
Layer type                   & Dense & Dense & Output observation               \\
Input dim.            & $1$  & $16$ & $1$                   \\
Output dim.            & $16$  & $1$ & $1$                   \\
Inc. bias                    & \checkmark & \checkmark & - \\
Weight prior $\sigma$        & $6.0 $ &$1.5$&-                 \\
Activation prior $\sigma$   & $10.0$   &$3.0$ &-                \\
Input obs. $\sigma$          & $0.02$ & - &-                  \\
Output obs. $\sigma$ & - &- & $0.05 $              \\
Dense recon. $\sigma$              & $5\times10^{-3}$ & $1\times10^{-2}$ & -                   \\
Activation function $g(\cdot)$ & Sigmoid & Linear & -  \\ \bottomrule                
\end{tabular}
\caption{The MLP-like factor graph used for the regression experiment.}
\label{tab:regression_model}
\end{table}

\section{Video Denoising Experiment Details}
\label{sec:app:video_denoising_experiment}
\setcounter{table}{0}
\renewcommand{\thetable}{\ref{sec:app:video_denoising_experiment}\arabic{table}}
\setcounter{figure}{0}
\renewcommand{\thefigure}{\ref{sec:app:video_denoising_experiment}\arabic{figure}}

\subsection{Single Layer Convolutional Factor Graph}
We use the transposed convolution model described in Table~\ref{tab:video_denoise_conv_model} for both per-frame and continual learning experiments. We run for $300$ GBP iterations on each frame with damping factor of $0.8$ and dropout factor of $0.6$ applied to the factor to variable messages. We used robust factor energies similar to the Tukey loss \citep{Tukey:1960}: quadratic within a Mahalanobis distance of $\robthresh$ from the mean, and flat outside.
\begin{table}[h!]
\centering
\begin{tabular}{lc}\toprule
Layer \#                     & 1                     \\ \midrule
Layer type                   & Transposed Conv.                 \\
Number of filters            & 4                     \\
Inc. bias                    & \checkmark \\
Kernel size                  & $3\times3$            \\
Conv. recon. $\sigma$              & 0.1                   \\
Recon $\robthresh$      & 1.4                  \\
Weight prior $\sigma$        & 0.018                 \\
Activation prior $\sigma$   & 0.5                   \\
Pixel obs. $\sigma$          & 0.2                  \\
Pixel obs $\robthresh$ & 0.2                  \\
Activation function $g(\cdot)$ & Linear     \\\bottomrule           
\end{tabular}
\caption{The single layer convolutional factor graph model used for the video denoising experiments. $\robthresh$ is the ``robust threshold'': the Mahalanobis distance beyond which the factor energy is flat rather than quadratic.}
\label{tab:video_denoise_conv_model}
\end{table}

\subsubsection{Preventing overfitting in single-layer model}
We find the single-layer convolutional model described above can overfit to the salt-and-pepper noise when the filters are learnt continually over the course of the video. We find that overfitting can be reduced by choosing the parameter prior at each frame to be an interpolation of the previous parameter posterior and the original parameter prior.

More concretely, for a parameter $\filter_{l,i}$ with original prior (before the first frame) $\mathcal{N}(\filter_{l,i};\mu_{\theta_l}, \sigma_{\theta_l}^2)$ and posterior from previous frames $\mathcal{N}(\filter_{l,i};\mu_{l,i}^{(t-1)}, (\sigma_{l,i}^{(t-1)})^2)$ we set its prior for a frame $t$ to be
\begin{align}
    p_t(\filter_{l,i}) &\leftarrow \mathcal{N}\lrb{\filter_{l,i};\mu_{l,i}, \lrb{\sigma_{l,i}}^2}~,\\
    \intertext{where}
    \mu_{l,i} &= \alpha \cdot \mu_{\theta_l} + (1 - \alpha) \cdot \mu_{l,i}^{(t-1)} \\
    \sigma_{l,i} &= \alpha \cdot \sigma_{\theta_l} + (1 - \alpha) \cdot \sigma_{l,i}^{(t-1)}~.
\end{align}
We used $\alpha=0.5$ for the single-layer, continual learning video denoising experiment.

\subsection{Five Layer Convolutional Factor Graph}
We use the transpose convolution model described in Table~\ref{tab:video_denoise_conv_model_five_layer} for both per-frame and continual learning experiments. We run for $500$ GBP iterations on each frame with damping factor of $0.8$ and dropout factor of $0.6$ applied to the factor to variable messages. We used robust factor energies similar to the Tukey loss \citep{Tukey:1960}: quadratic within a Mahalanobis distance of $\robthresh$ from the mean, and flat outside.

To increase the spatial extent of the activations, we use upsampling layers within which each output $x_l^{(a,b,c)}$ connects to a $\filtsize\times \filtsize$ patch of the input $\inputpatch_{l-1}^{(a,b,c)}$ via a factor with energy \myeqref{eq:Eupsample}.

\begin{table}[h!]
\centering
\begin{tabular}{lccccc}\toprule
Layer \#                     & 1  & 2 & 3 & 4 & 5                   \\ \midrule
Layer type                   & Transposed Conv. & Upsample &   Transposed Conv. & Upsample & Transposed Conv.               \\
Number of filters            & 4 &- & 8 & - & 8                    \\
Inc. bias                    & \checkmark & - & \checkmark & N/A & \checkmark \\
Kernel size                  & $3\times3$ & $2\times2$  & $3\times3$ & $2\times2$ & $3\times3$ \\
Conv. recon. $\sigma$              & 0.12 & 0.03 & 0.07  & 0.03 & 0.07  \\
Recon $\robthresh$      & 2.5 & - &  - & - & -     \\
Weight prior $\sigma$        & 0.15 & -  & 0.3 & - & 0.3  \\
Activation prior $\sigma$   & 0.5 & -  & 0.5 & - & 0.5 \\
Pixel obs. $\sigma$          & 0.2   & -& -& -& -               \\
Pixel obs $\robthresh$ & 0.2& -& -& -& -                  \\
Activation func. $g(\cdot)$ & Linear & - & Leaky ReLU & - & Leaky ReLU    \\   \bottomrule        
\end{tabular}
\caption{The five-layer convolutional factor graph model used for the video denoising experiments. $\robthresh$ is the ``robust threshold'': the Mahalanobis distance beyond which the factor energy is flat rather than quadratic.}
\label{tab:video_denoise_conv_model_five_layer}
\end{table}

\FloatBarrier

\subsection{Pairwise Factor Graph}
The factor hyperparameters for the pairwise smoother baseline are given in Table~\ref{tab:video_denoise_baseline_model}. We denoise by running $200$ GBP iterations on each frame with damping factor of $0.7$ applied to the factor to variable messages. We used robust factor energies similar to the Tukey loss \citep{Tukey:1960}: quadratic with a Mahalanobis distance of $\robthresh$ from the mean, and flat outside.
\begin{table}[h!]
\centering
\begin{tabular}{lc}\toprule
Layer \#                     & 1                  \\ \midrule
Layer type                   & Pairwise smoothing \\
Pixel obs. $\sigma$          & 0.2                \\
Pixel obs $\robthresh$  & 0.14               \\
Pairwise $\sigma$            & 1.3               \\
Pairwise $\robthresh$   & 0.35              \\\bottomrule
\end{tabular}
\caption{The pairwise smoothing baseline model used for the video denoising experiments. $\robthresh$ is the ``robust threshold'': the Mahalanobis distance beyond which the factor energy is flat rather than quadratic.}
\label{tab:video_denoise_baseline_model}
\end{table}

\newpage
\section{Denoised Video Frame Example}
\label{sec:app:denoised_images}
\FloatBarrier
\begin{figure}[h!]
    \centering
    \begin{subfigure}{0.44\textwidth}
        \centering
        \includegraphics[width=\textwidth]{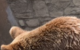}
        \caption{Clean image}
        \label{subfig:frame_clean}
    \end{subfigure}\hfill
    \begin{subfigure}{0.44\textwidth}
        \centering
        \includegraphics[width=\textwidth]{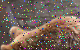}
        \caption{Corrupted image}
        \label{subfig:frame_noisy}
    \end{subfigure} \\
    \begin{subfigure}{0.44\textwidth}
    \centering
        \includegraphics[width=\textwidth]{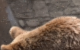}
        \caption{Per-frame learning, single layer}
        \label{subfig:frame_per_frame}
    \end{subfigure}\hfill
    \begin{subfigure}{0.44\textwidth}
    \centering
        \includegraphics[width=\textwidth]{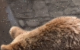}
        \caption{Continual learning, single layer}
        \label{subfig:frame_continual}
    \end{subfigure}\\
    \begin{subfigure}{0.44\textwidth}
    \centering
        \includegraphics[width=\textwidth]{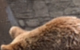}
        \caption{Per-frame learning, five layer}
        \label{subfig:frame_per_frame_five_layer}
    \end{subfigure}\hfill
    \begin{subfigure}{0.44\textwidth}
    \centering
        \includegraphics[width=\textwidth]{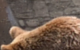}
        \caption{Continual learning, five layer}
        \label{subfig:frame_continual_five_layer}
    \end{subfigure}\\
    \begin{subfigure}{0.44\textwidth}
        \includegraphics[width=\textwidth]{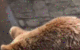}
        \caption{Pairwise smoothing}
        \label{subfig:frame_pairwise}
    \end{subfigure}
    \caption{{\bf A crop from frame 5.} The learnt models are able to remove more noise while retaining more high-frequency signal.}
    \label{fig:video_denoising_frames_crop}
\end{figure}

\begin{figure}[h!]
    \centering
    \begin{subfigure}{0.48\textwidth}
        \centering
        \includegraphics[width=\textwidth]{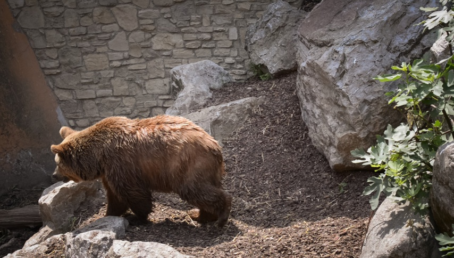}
        \caption{Clean image}
        \label{subfig:frame_clean}
    \end{subfigure}\hfill
    \begin{subfigure}{0.48\textwidth}
        \centering
        \includegraphics[width=\textwidth]{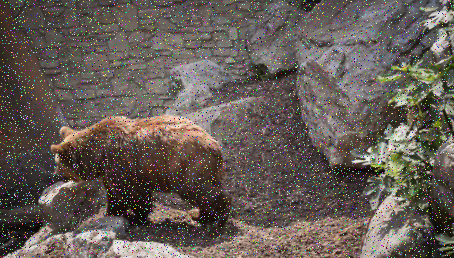}
        \caption{Corrupted image}
        \label{subfig:frame_noisy}
    \end{subfigure} \\
    \begin{subfigure}{0.48\textwidth}
    \centering
        \includegraphics[width=\textwidth]{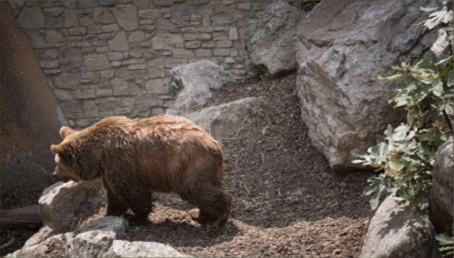}
        \caption{Per-frame learning, single layer, $\psnr=32.6$}
        \label{subfig:frame_per_frame}
    \end{subfigure}\hfill
    \begin{subfigure}{0.48\textwidth}
    \centering
        \includegraphics[width=\textwidth]{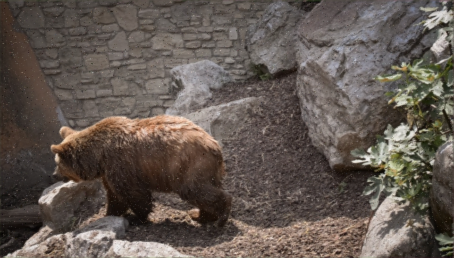}
        \caption{Continual learning, single layer, $\psnr=32.9$}
        \label{subfig:frame_continual}
    \end{subfigure}\\
    \begin{subfigure}{0.48\textwidth}
    \centering
        \includegraphics[width=\textwidth]{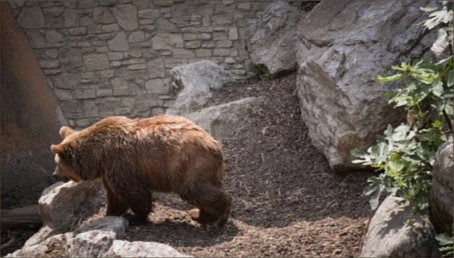}
        \caption{Per-frame learning, five layer, $\psnr=33.4$}
        \label{subfig:frame_per_frame_five_layer}
    \end{subfigure}\hfill
    \begin{subfigure}{0.48\textwidth}
    \centering
        \includegraphics[width=\textwidth]{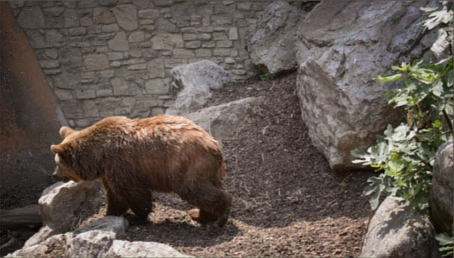}
        \caption{Continual learning, five layer, $\psnr=34.3$}
        \label{subfig:frame_continual_five_layer}
    \end{subfigure}\\
    \begin{subfigure}{0.48\textwidth}
        \includegraphics[width=\textwidth]{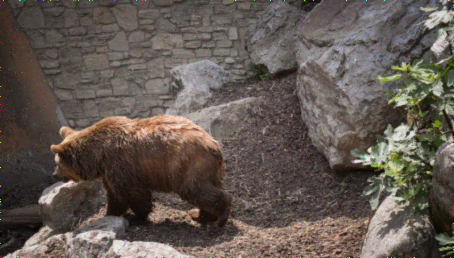}
        \caption{Pairwise smoothing, $\psnr=31.1$}
        \label{subfig:frame_pairwise}
    \end{subfigure}
    \caption{{\bf Frame 5 denoised by each method.}}
    \label{fig:video_denoising_frames}
\end{figure}

\FloatBarrier

\section{MNIST experiment details}
\label{sec:app:mnist_experiment}

\subsection{Convolutional Factor Graph}
\FloatBarrier
\label{sec:app:mnist_model}
For the MNIST experiment we tuned the factor graph architecture and the other hyperparameters on a validation set of $9,000$ examples sampled uniformly from the training set. We then used the same tuned model for every training set size.

The final convolutional factor graph model is summarised in Table~\ref{tab:mnist_factor_graph_config}. Pixel variables are fixed at their observed values. To produce the results shown in Fig.~\ref{fig:mnist}, we train with continual learning, with a batchsize of $50$ and run $500$ GBP iterations on each batch. At test time, we fix the parameters and run GBP for $300$ iterations per test batch of $200$ examples. We apply a damping factor of $0.9$ and a dropout factor of $0.5$ to the factor to variable messages at both train and test time. Note that each iteration includes the updating of the messages in each layer, sweeping from image layer to classification head, and then sweeping back to image. 

\begin{table}[h!]
\centering
\small{
\begin{tabular}{lcccc}\toprule
Layer \#                     & 1                         & 2          & 3                         & 4$^\dagger$ \\ \midrule
Layer type                   & Conv.                     & Max pool  & Dense & Softmax \\
Num. filters            & 16                         & -        & -                        & -  \\
Kernel size                  & $5\times5$                & $2\times2$ & - & -     \\
Dense num. inputs       & -                       & -        & 2304 & - \\
Dense num. outputs      & -                      & -      & 10                       & -\\
Inc. bias                    & \checkmark & -       & \checkmark & - \\
Weight prior $\sigma$        & 0.1                        & -        & 0.15 &-     \\
Activation prior $\sigma$   & 3.0                        & 3.0         & 2.0                        & - \\
Recon. $\sigma$              & 0.01                       & 0.01       & 0.01 & - \\
Activation func. $g(\cdot)$                  & Leaky ReLU                       & Linear       & Linear                       &  -     \\
Class observation $\sigma$   & -                      & -        & -                      &  0.01 \\\bottomrule
\end{tabular}}
\caption{The convolutional factor graph model used for the MNIST experiment. The dimensions of the layers are chosen such that no padding is necessary. Note that ``Recon $\sigma$'' denotes the strength of the factors which connect one layer to the next. $^\dagger$Softmax layer only included at training time, when class observation is available. As test time, the final state of the last layer variables after GBP inference is treated as the logit prediction.}
\label{tab:mnist_factor_graph_config}
\end{table}
\FloatBarrier

\subsection{Linear Classifier Baseline}
\label{sec:app:mnist_baselines}
\FloatBarrier
As a baseline we used a linear classifier trained with Adam to minimise the cross-entropy loss. We tuned the step size and number of epochs on a randomly sampled validation set comprising $9,000$ training set examples. We tuned a separate linear classifier configuration for each training set size. As with the factor graph model, we used a batchsize of $50$ for all experiments in Fig.~\ref{fig:mnist}.
\FloatBarrier

\subsection{CNN + Replay Buffer Baseline}
\FloatBarrier
We train a CNN with Adam as a comparison against our convolutional factor graph trained with GBP Learning. To make this comparison as fair as possible we
\begin{itemize}
	\item Use a CNN architecture as close as possible to our factor graph (Table~\ref{tab:mnist_factor_graph_config}). The CNN architecture is summarised in Table~\ref{tab:mnist_cnn_config}.
	\item Train for only one epoch, but equip the CNN with a FIFO replay buffer to reduce forgetting. A fixed number of elements from each batch are randomly selected and added to the buffer. The size of the buffer necessary to match the performance of GBP Learning then provides an indication as to the efficacy of our continual learning approach. We evaluate buffers of various sizes. To allow comparison across different dataset sizes, we express buffer size as a fraction of the training set size. 
\end{itemize}

We tune the following hyperparameters of the CNN + replay buffer method, finding a different configuration \emph{for each combination of buffer size} (as fraction of training set size) \emph{and training set size}:
\begin{itemize}
	\item Number of elements in each batch added to the buffer
	\item Number of steps to take on each training set batch
	\item Number of steps to take on batches sampled from the replay buffer, for each training set batch
	\item Step size for training set batch updates
	\item Step size for replay buffer updates.
\end{itemize}
These hyperparameters were selected based on a validation set of $9,000$ examples, sampled uniformly from the training set. The Adam optimiser \cite{kingma2014adam} was used for all parameter updates.

\begin{table}[h!]
\centering
\small{
\begin{tabular}{lccc}\toprule
Layer \#                     & 1                         & 2          & 3                       \\ \midrule
Layer type                   & Conv.                     & Max pool  & Dense \\
Num. filters            & 16                         & -        & -                         \\
Kernel size                  & $5\times5$                & $2\times2$ & -      \\
Dense num. inputs       & -                       & -        & 2304  \\
Dense num. outputs      & -                      & -      & 10                       \\
Inc. bias                    & \checkmark & -       & \checkmark 
\\\bottomrule
\end{tabular}}
\caption{The baseline CNN architecture.}
\label{tab:mnist_cnn_config}
\end{table}
\FloatBarrier

\subsection{Asynchronous Training}
\label{sec:app:mnist_async}
\FloatBarrier
To test the robustness of our method to asynchronous training regimes, we evaluated training the model in Table~\ref{tab:mnist_factor_graph_config} on MNIST using random layer schedules. At each iteration, we uniformly sample $4$ integers in the interval $[1,4]$ with replacement. These integers index the different layers of the network. We then update the messages within each of the sampled layers, in the sampled order. Note that we sample with replacement, meaning that some layers may not be updated at all during a given iteration, and some may be updated multiple times.

\begin{figure}[t]
    \centering
    \includegraphics[width=0.7\textwidth]{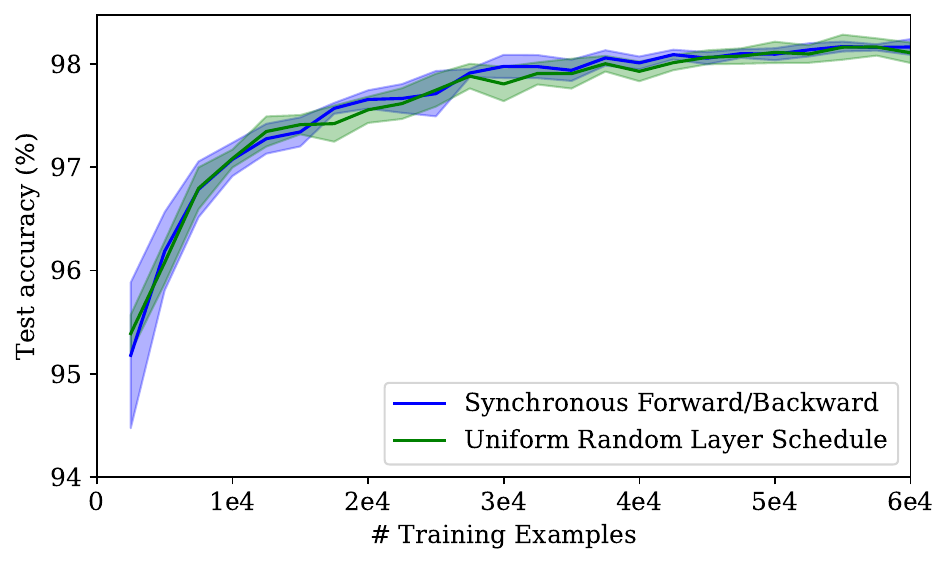}
    \caption{The evolution in test accuracy over the course of training, for a convolutional factor graph (Table~\ref{tab:mnist_factor_graph_config}) trained on MNIST, with different message schedules: i) synchronous forward/backward sweeps and ii) random layer ordering. Intervals represent $\pm1$ standard error either side of the mean, over $5$ random seeds.}
    \label{fig:async_training}
\end{figure}

The progression of test accuracy for both asynchronous and synchronous training is presented in Fig.~\ref{fig:async_training}. Both regimes exhibit similar performance throughout training, suggesting that GBP Learning can be executed in a distributed and asynchronous manner with little loss in performance.

\FloatBarrier

\subsection{Dependence on number of iterations}
\label{sec:app:mnist_n_iter_char}
\FloatBarrier
We seek to understand how the performance of our models depends on different levels of compute at train and test time. We train replicas of the convolutional factor graph model (summarised in Table~\ref{tab:mnist_factor_graph_config}) on MNIST, each with a different number of GBP iterations per training batch. We then evaluate the test accuracy of each of the trained models multiple times: for differing numbers of test time GBP iterations. All configurations used a batch size of $50$ at train time and $200$ at test time. We ran our experiments with $50, 100, 200, 400, 800, 1600$ iterations per batch at both train time and test time.

\begin{figure}[t]
    \centering
    \begin{subfigure}{0.48\textwidth}
        \includegraphics[width=\textwidth]{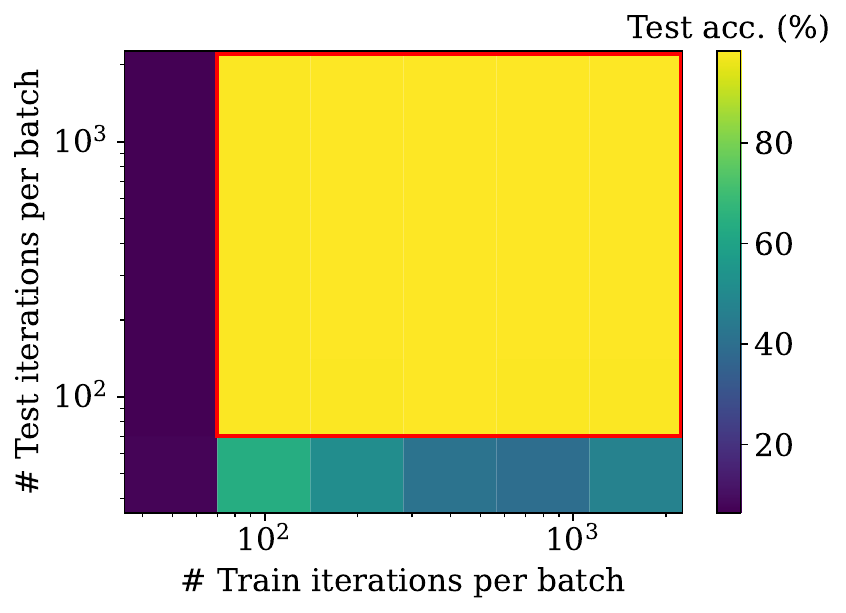}
    \end{subfigure}
    \begin{subfigure}{0.48\textwidth}
        \includegraphics[width=\textwidth]{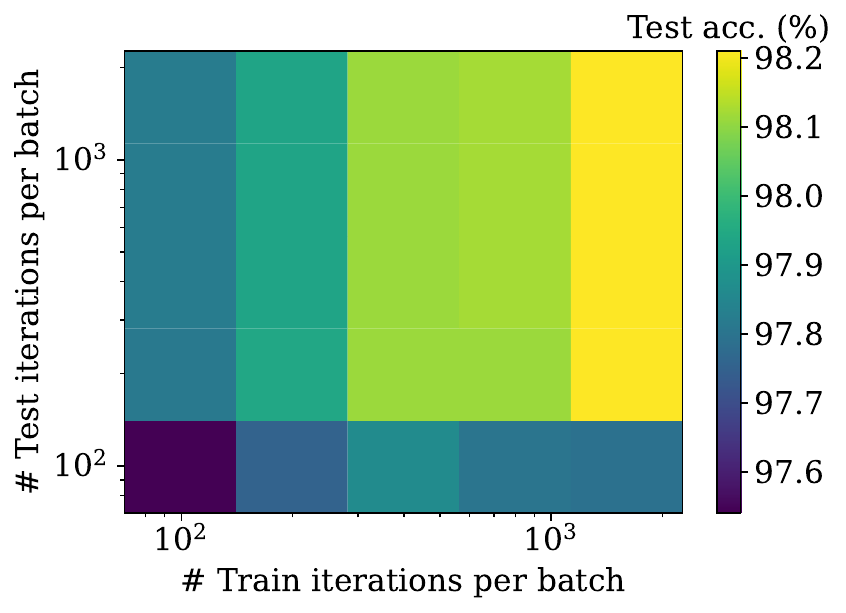}
    \end{subfigure}
    \caption{Dependence of MNIST test accuracy on the number of GBP iterations at train and test time. The right-hand plot covers the range of iterations marked by the red box in the left-hand plot, with the colours rescaled accordingly.}
    \label{fig:mnist_n_iter_char}
\end{figure}

The results are presented in Fig.~\ref{fig:mnist_n_iter_char}. They show that good test time performance can be achieved with relatively few GBP iterations per training batch ($\sim200$), as long as a sufficient number of iterations is run at test time ($\geqslant200$). However, the best performance is achieved with $1600$ iterations per train batch.

\FloatBarrier
\end{document}